\documentclass[]{eccv}

\usepackage{xcolor}
\usepackage{graphicx}
\usepackage{amsmath}
\usepackage{amssymb}
\usepackage{booktabs}
\usepackage{xspace}
\usepackage{pifont}%
\usepackage{bbm}
\usepackage{multirow}
\usepackage{calc}
\usepackage{cite}
\usepackage{orcidlink}
\usepackage{dcolumn}
\usepackage{tikz}
\usetikzlibrary{positioning}
\usetikzlibrary{calc}

\begin{document}
\pagestyle{headings}
\mainmatter
\def\ECCVSubNumber{1417}

\title{ViewFormer: NeRF-free Neural Rendering \\ from Few Images Using Transformers}

\titlerunning{ViewFormer: NeRF-free Neural Rendering from Few Images}
\author{Jonáš Kulhánek \inst{1,2}\orcidlink{0000-0002-8437-3626} \and
Erik Derner\inst{1}\orcidlink{0000-0002-7588-7668} \and
Torsten Sattler\inst{1}\orcidlink{0000-0001-9760-4553} \and
Robert Babuška\inst{1,3}\orcidlink{0000-0001-9578-8598}}
\authorrunning{J. Kulhánek et al.}
\institute{Czech Institute of Informatics, Robotics and Cybernetics,\\
Czech Technical University in Prague \and
Faculty of Electrical Engineering, Czech Technical University in Prague \and
Cognitive Robotics, Faculty of 3mE, Delft University of Technology
}
\maketitle

\begin{abstract}
Novel view synthesis is a long-standing problem.
 In this work, we consider a variant of the problem where we are given only a few context views sparsely covering a scene or an object. 
The goal is to predict novel viewpoints in the scene, which requires learning priors.
The current state of the art is based on Neural Radiance Field (NeRF), and while achieving impressive results, the methods suffer from long training times as they require evaluating millions of 3D point samples via a neural network for each image. 
We propose a 2D-only method that maps multiple context views and a query pose to a new image in a single pass of a neural network.
Our model uses a two-stage architecture consisting of a codebook and a transformer model. The codebook is used to embed individual images into a smaller latent space, and the transformer solves the view synthesis task in this more compact space.
To train our model efficiently, we introduce a novel \emph{branching attention} mechanism that allows us to use the same model not only for neural rendering but also for camera pose estimation. 
Experimental results on real-world scenes show that our approach is competitive compared to NeRF-based methods while not reasoning explicitly in 3D, and it is faster to train. 
\end{abstract}

\section{Introduction}\label{sec:introduction}
Image-based novel view synthesis, \ie, rendering a 3D scene from a novel viewpoint given a set of context views (images and camera poses), is a long-standing problem in computer graphics with applications ranging from robotics (\eg planning to grasp objects) to augmented and virtual reality (\eg interactive virtual meetings).
Recently, the field has gained a lot of popularity thanks to Neural Radiance Field (NeRF) methods \cite{mildenhall2020nerf,barron2021mip} that were successfully applied to the problem and outperformed prior approaches.
We distinguish between two variants of the view synthesis problem. The first variant renders a novel view from multiple context images taken from similar viewpoints~\cite{mildenhall2020nerf,wang2021ibrnet}. 
Only a (very) sparse set of context images is provided in the second variant~\cite{yu2021pixelnerf,reizenstein2021common}, \ie, larger viewpoint variations and missing observations need to be handled. 
The latter task is much more difficult as it is necessary to learn suitable priors that can be used to predict  unseen scene parts. 
\begin{figure}[t]
\centering
\newcommand{\sz}{0.24\linewidth}
\begin{tikzpicture}[
scale=0.8,
 image/.style = {text width=1.6cm, anchor=north west,
                 inner sep=0pt, outer sep=0pt},
 simage/.style = {text width=0.8cm-0.4pt, anchor=north west,
                 inner sep=0pt, outer sep=0pt},
label/.style = { minimum height=0.4cm },
node distance = 1pt and 1pt
                        ] 
\scriptsize
\begin{scope}[]
\node [simage] at (0, 0)
{\includegraphics[width=\linewidth]{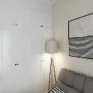}};
\node [simage] at (1cm+0.5pt, 0)
{\includegraphics[width=\linewidth]{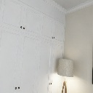}};
\node [simage] at (0, -1cm-0.5pt)
{\includegraphics[width=\linewidth]{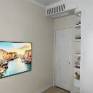}};
\node [simage] at (1cm+0.5pt, -1cm-0.5pt)
{\includegraphics[width=\linewidth]{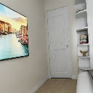}};
\node [simage] at (-2cm-1.0pt, 0)
{\includegraphics[width=\linewidth]{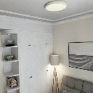}};
\node [simage] at (-1cm-0.5pt, 0)
{\includegraphics[width=\linewidth]{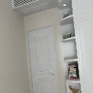}};
\node [simage] at (-2cm-1.0pt, -1cm-0.5pt)
{\includegraphics[width=\linewidth]{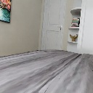}};
\node [simage] at (-1cm-0.5pt, -1cm-0.5pt)
{\includegraphics[width=\linewidth]{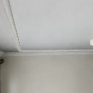}};
\node [image] (gt) at (2cm+1pt, 0)
{\includegraphics[width=\linewidth] {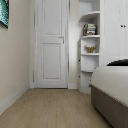}};
\node [image] (gen) at (4cm+2pt, 0)
{\includegraphics[width=\linewidth]{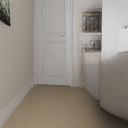}};
\end{scope}
\begin{scope}[yshift=-2cm-1pt-1mm]
\node [simage] at (0, 0)
{\includegraphics[width=\linewidth]{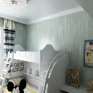}};
\node [simage] at (1cm+0.5pt, 0)
{\includegraphics[width=\linewidth]{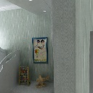}};
\node [simage] at (0, -1cm-0.5pt)
{\includegraphics[width=\linewidth]{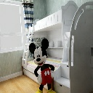}};
\node [simage] at (1cm+0.5pt, -1cm-0.5pt)
{\includegraphics[width=\linewidth]{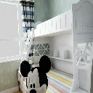}};
\node [simage] at (-2cm-1pt, 0)
{\includegraphics[width=\linewidth]{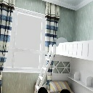}};
\node [simage] at (-1cm-0.5pt, 0)
{\includegraphics[width=\linewidth]{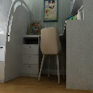}};
\node [simage] at (-2cm-1pt, -1cm-0.5pt)
{\includegraphics[width=\linewidth]{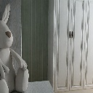}};
\node [simage] at (-1cm-0.5pt, -1cm-0.5pt)
{\includegraphics[width=\linewidth]{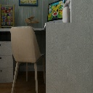}};
\node [image] at (2cm+1pt, 0)
{\includegraphics[width=\linewidth] {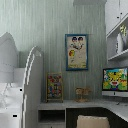}};
\node [image] at (4cm+2pt, 0)
{\includegraphics[width=\linewidth]{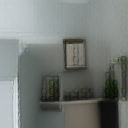}};
\end{scope}

\begin{scope}[xshift=62mm]
\node [simage] at (0, 0)
{\includegraphics[width=\linewidth]{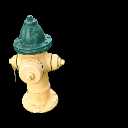}};
\node [simage] at (1cm+0.5pt, 0)
{\includegraphics[width=\linewidth]{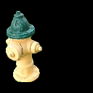}};
\node [simage] at (0, -1cm-0.5pt)
{\includegraphics[width=\linewidth]{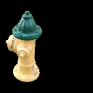}};
\node [simage] at (1cm+0.5pt, -1cm-0.5pt)
{\includegraphics[width=\linewidth]{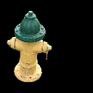}};
\node [image] (gt2) at (2cm+1pt, 0)
{\includegraphics[width=\linewidth]{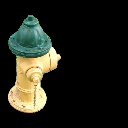}};
\node [image] (gen2) at (4cm+2pt, 0)
{\includegraphics[width=\linewidth] {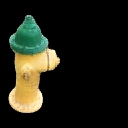}};
\end{scope}
\begin{scope}[xshift=62mm,yshift=-2cm-1pt-1mm]
\node [image] at (0, 0)
{\includegraphics[width=\linewidth]{resources/teaser/hydrant/00.png}};
\node [image] at (2cm+1pt, 0)
{\includegraphics[width=\linewidth]{resources/teaser/hydrant/00000002-gt.png}};
\node [image] at (4cm+2pt, 0)
{\includegraphics[width=\linewidth]{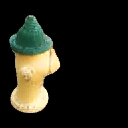}};
\end{scope}

\node[label] at (0, 6pt) { context images };
\node [label,above=-0.5pt of gt] {GT};
\node [label,above=-1pt of gen] {generated};
\begin{scope}[xshift=62mm]
\node[label] at (28pt, 6pt) { context images };
\node [label,above=-0.5pt of gt2] {GT};
\node [label,above=-1pt of gen2] {generated};
\end{scope}
\end{tikzpicture}

\caption{
Our novel view synthesis method renders images of previously unseen objects based on a few context images. It operates in 2D space without any explicit 3D reasoning (as opposed to NeRF-based approaches \cite{yu2021pixelnerf,reizenstein2021common}). The results are shown on the CO3D  \cite{reizenstein2021common} (\textbf{right}) and InteriorNet \cite{li2018interiornet} (\textbf{left}) datasets rendered for unseen scenes
\label{fig:teaser}
}
\end{figure}
This paper focuses on the second variant.

Recently, generalizable NeRF-based approaches have been proposed to tackle this problem by learning priors for a class of objects and scenes~\cite{yu2021pixelnerf,reizenstein2021common}.
Instead of learning a radiance field for each scene, they use context views captured from the target scene to construct the radiance field on the fly by projecting the image features from all context views into 3D.
Highly optimized NeRF approaches \cite{reiser2021kilonerf,muller2022instant,garbin2021fastnerf,yu2021plenoctrees} can be sped up by tuning or caching the radiance field representation \cite{muller2022instant}, although often requiring lots of images per scene.
To the best of our knowledge, these techniques do not apply to generalizable NeRF-based methods that do not learn a scene-specific radiance field, and take thousands of GPU-hours to train \cite{reizenstein2021common}.
In contrast, 2D-only feed-forward networks can be highly efficient. 
However, explicitly encoding 3D geometric principles in them can be challenging. 
In our work, we thus pose the question: \emph{Is reasoning in 3D necessary for high-quality novel view synthesis, or can a purely image-based method achieve a competitive performance?}

Recently, Rombach \etal \cite{rombach2021geometry} successfully tackled single-view novel view synthesis, where the model was able to predict novel views without explicit 3D reasoning. Inspired by these findings, we tackle the more complex problem of multi-view novel view synthesis. To answer the question, we propose a method with no explicit 3D reasoning able to predict novel views using multiple context images in a forward pass of a neural network.
We train our model on a large collection of diverse scenes to enable the model to learn 3D priors implicitly.
Our approach is able to render a view in a novel scene, unseen at training time, three orders of magnitude faster than state-of-the-art (SoTA) NeRF-based approaches \cite{reizenstein2021common}, while also being ten times faster to train.
Furthermore, we are able to train a single model to render multiple classes of scenes (see \cref{fig:teaser}), whereas the SoTA NeRF-based approaches typically train per-class models \cite{reizenstein2021common}. 

Our model uses a two-stage architecture consisting of a Vector Quantized-Variational Autoencoder (VQ-VAE) codebook \cite{oord2017neural} and a transformer model.
The codebook model is used to embed individual images into a smaller latent space. 
The transformer solves the novel view synthesis task in this latent space before the image is recovered via a decoder. This enables the codebook to focus on finer details in images while the transformer operates on shorter input sequences,
reducing the quadratic memory complexity of its attention layer.

For training, we pass a sequence of views into the transformer and optimize it for all context sizes at the same time, effectively utilizing all images in the training batch, which is different from other methods \cite{eslami2018neural,parmar2018image,esser2021taming,ramesh2021zero} that train only one query view.
Unlike autoregressive models \cite{parmar2018image,esser2021taming,ramesh2021zero}, we do not decode images token-by-token but all tokens are decoded at once which is both faster and mathematically exact (while autoregressive models rely on greedy strategies).
Our approach can be considered a combination of autoregressive \cite{vaswani2017attention,radford2019language} and masked \cite{devlin2019bert} transformer models.
With the standard attention mechanism, the complexity would be quadratic in the number of views, because we would have to stack different query views corresponding to different context sizes along the batch dimension.
Therefore, we propose a novel attention mechanism called \emph{branching attention} with constant overhead regardless of how many query views we optimize.
Our attention mechanism also allows us to optimize the same model for the camera pose estimation task  -- predicting the query image's camera pose given a set of context views. Since this task can be considered an ``inverse" of the novel view synthesis task~\cite{yen2020inerf}, we consider the ability to perform both tasks via the same model to be an intriguing property. Even though the localization results are not yet competitive with state-of-the-art localization pipelines, we achieve a similar level of pose accuracy as comparable methods such as~\cite{shavit2021learning,balntas2018relocnet}.

In summary, this paper makes the following contributions: 
\textbf{1}) We propose an efficient novel view synthesis approach that does not use explicit 3D reasoning. 
Our two-stage method consisting of a codebook model and a transformer is competitive with state-of-the-art NeRF-based approaches while being more efficient to train.
Compared to similar methods that do not use explicit 3D reasoning~\cite{eslami2018neural,tobin2019geometry,chen2021str}, our approach is not only evaluated on synthetic data but performs well on real-world scenes.
\textbf{2}) Our transformer model is a combination of an autoregressive and a masked transformer. We propose a novel attention mechanism called \emph{branching attention} that allows us to optimize for multiple context sizes at once with a constant memory overhead. 
\textbf{3}) Thanks to the branching attention, our model can both render a novel view from a given pose and predict the pose for a given image.
\textbf{4}) Our source code and pre-trained models are publicly available at \url{https://github.com/jkulhanek/viewformer}.

\section{Related work}
\PAR{Novel view synthesis} 
has a long history~\cite{shum2000review,chan2007image}.
Recently, deep learning techniques have been applied with great success, enabling higher realism \cite{choi2019extreme,hedman2018deep,riegler2020free,riegler2021stable,martin2021nerf}.
Some approaches use explicit reconstructed geometry to warp context images into the target view \cite{choi2019extreme,hedman2018deep,thies2020image,riegler2020free,riegler2021stable}.
In our approach, we do not require any proxy geometry and only operate on 2D images.

Neural Radiance Field methods
\cite{mildenhall2020nerf,liu2020neural,jain2021putting,barron2021mip,yu2021plenoctrees,reiser2021kilonerf,martin2021nerf} use neural networks to represent the continuous volumetric scene function.
To render a view, for each pixel in the image plane, they project a ray into 3D space and query the radiance field in 3D points along each ray. The radiance field is trained for each scene separately.
Some methods generalize to new scenes by conditioning the continuous volumetric function on the context images \cite{saito2019pifu,sitzmann2019scene}, which allows them to utilize trained priors and render views from scenes on which the model was not trained, much like our approach.
Other approaches remove the trainable continuous volumetric scene function altogether. 
Instead, they reproject the context image's features into the 3D space and apply the NeRF-based rendering pipeline on top of this representation \cite{yu2021pixelnerf,reizenstein2021common,trevithick2021grf,henzler2021unsupervised,wang2021ibrnet}.
Similarly to these methods, our approach also utilizes few context views
(less than 20), and it also generalizes to unseen objects. However, we do not use the continuous volumetric function nor the reprojection into the 3D space.
A different approach, IBRNet \cite{wang2021ibrnet}, learns to copy existing colors from context views, effectively interpolating the context views. 
Unlike ours, it thus cannot be applied to the settings where the object is not covered enough by the context views \cite{yu2021pixelnerf,reizenstein2021common,trevithick2021grf,henzler2021unsupervised}.

A different line of work directly maps 2D context images to the 2D query image using an end-to-end neural network \cite{eslami2018neural,tobin2019geometry,chen2021str}.
GQN-based methods \cite{eslami2018neural,tobin2019geometry,chen2021str} apply a CNN to context images and camera poses and combine the resulting features. While some GQN methods \cite{eslami2018neural,chen2021str} do not use any explicit 3D reasoning (same as our approach), Tobin \etal \cite{tobin2019geometry} uses an epipolar attention to aggregate the features from the context views.
We optimize our model on all context images and fully utilize the training sequences, whereas GQN methods optimize only a single query view.

A recent work by Rombach \etal \cite{rombach2021geometry} proposed an approach for novel view synthesis without explicit 3D modeling. They used a codebook and a transformer model to map a single context view to a novel view from a different pose.
Their approach is limited in its scope to mostly forward-facing scenes where it is easier to render the novel view given a single context view and the poses have to be close to one another.
It cannot be extended to more views due to the limit on the sequence size of the transformer model. 
In contrast, in our approach, we focus on using multiple context views, which we tackle through
the proposed branching attention. 
Furthermore, we can jointly train the same model for both the novel view synthesis and camera pose estimation and our decoding is faster because we decode the output at once instead of autoregressive decoding.

\PAR{Visual localization.}
There is an enormous body of work tackling the problem of localization, where the goal is to output the camera pose given the camera image.
\emph{Structure-based} approaches use correspondences between 2D pixel positions and 3D scene coordinates for camera pose estimation \cite{Brachmann2021TPAMI,sattler2016efficient,sarlin2019cvpr,Shotton2013CVPR,Li2012ECCV,Cavallari2019TPAMI,Lynen2020IJRR}. Our method does not explicitly reason in 3D space, and the camera pose is instead predicted by the network.
Simple \emph{image retrieval} (IR) approaches store a database of all images with camera poses and for each query image they try to find the most similar images \cite{sattler2019understanding,camposeco2018hybrid,derner2021change,zhang2006image,cao2013graph,irschara2009structure} and use them to estimate the pose of the query. 
IR methods can also be used to select relevant images for accurate pose estimation \cite{bhayani2021iccv,zhang2006image,zhou2020icra,sarlin2019cvpr,irschara2009structure}.

\emph{Pose regression} methods 
train a convolutional neural network (CNN) to regress the camera pose of an input image. There are two categories: \emph{absolute pose regression} (APR) methods \cite{kendall2015posenet,brahmbhatt2018geometry,kendall2016modelling,blanton2020extending,chen2021direct,moreau2021lens,li2021transcampgt,shavit2021learning} and \emph{relative pose regression} (RPR) methods \cite{balntas2018relocnet,laskar2017camera,melekhov2017relative,ding2019camnet,li2021transcampgt}.
It was shown \cite{sattler2019understanding} that APR is often not (much) more accurate than IR.
RPR methods do not train a CNN per scene or a set of scenes, but instead, condition the CNN on a set of context views.
While our approach performs relative pose regression, the main focus of our method is on the novel view synthesis. 
Some pose regression methods use novel view synthesis methods \cite{moreau2021lens,ng2021reassessing,chen2021direct,mueller2019image},
however, they assume there is a method that generates images, whereas our method performs both the novel view synthesis and camera pose regression in a single model. 
\emph{Iterative refinement} pose regression methods \cite{yen2020inerf,sarlin2021back} start with an initial camera pose estimate and refine it by an iterative process,
however, our approach generates novel views and the camera pose estimates in a single forward pass.

\section{Method}\label{sec:method}
In this work, we tackle the problem of image-based novel view synthesis -- given a set of \emph{context} views, the algorithm has to generate the image it would most likely observe from a \emph{query} camera pose. We focus on the case where the number of context views is small, and the views sparsely cover the 3D scene. Thus, the algorithm must hallucinate parts of the scene in a manner consistent with the context views.
Therefore, it is necessary to learn a prior over a class of scenes (\eg, all indoor environments) and use this prior for novel scenes. Besides rendering novel views, our model can also perform camera pose estimation, \ie, the ``inverse" of the view synthesis task: given a set of context views and a query image, the model outputs the camera pose from which the image was taken.

\begin{figure}[t!]
\centering
\includegraphics[width=0.9\linewidth, trim={0 0.2cm 0 0},clip]{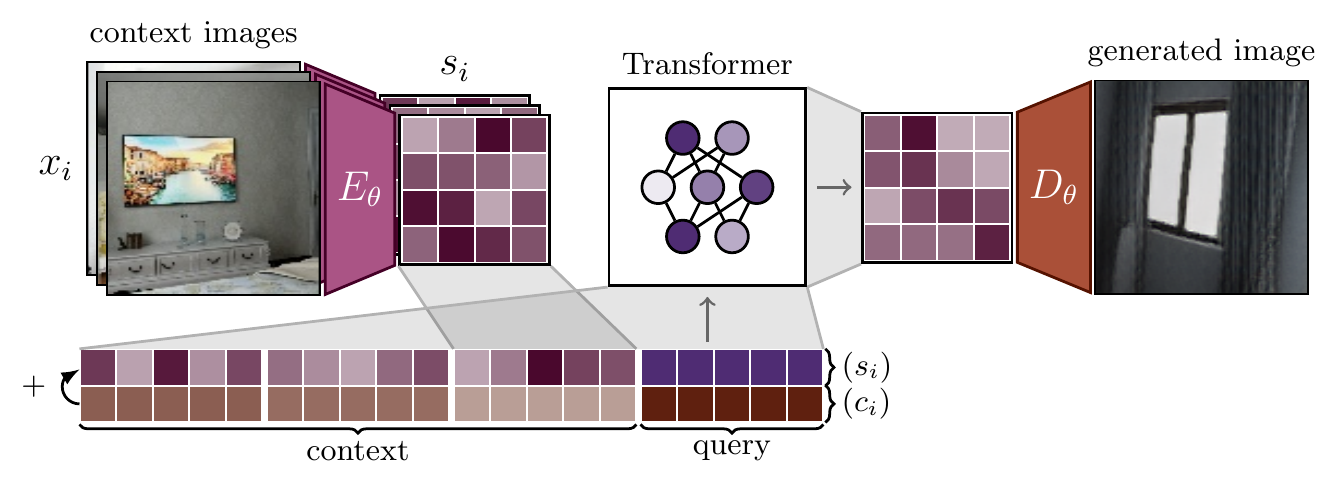}
\caption{Inference pipeline. The context images $x_i$ are encoded by the codebook's encoder $E_\theta$ to the code representation $s_i$. We embed all tokens in $s_i$, and add the transformed camera pose $c_i$. 
The transformer generates the image tokens which are decoded by the codebook's decoder $D_\theta$\label{fig:pipeline}}
\end{figure}

Our framework consists of two components: a codebook model and a transformer model. The codebook is used to map images to a smaller discrete latent space (\emph{code space}), and back to the image space. In the \emph{code space}, each image is represented by a sequence of \emph{tokens}. For the novel view synthesis task, the transformer is given a set of context views in the code space and the query camera pose, and it generates an image in the \emph{code space}. The codebook then maps the image tokens back to the image space. See \cref{fig:pipeline} for an overview. For the camera pose estimation task, the transformer is given the set of context views and the query image in the code space, and it generates the camera pose using a regression head attached to the output of the transformer corresponding to the query image tokens.

Having the codebook and the transformer as separate components was inspired by the recent work on image generation \cite{esser2021taming,ramesh2021zero,rombach2021geometry}.  
The main motivation is to decrease it sequence size, because the required memory grows quadratically with it. It also allows us to separate image generation and view synthesis, enabling us to train the transformer 
more efficiently in a simpler space. 

\PAR{Codebook model}is a VQ-VAE \cite{oord2017neural,razavi2019generating}, which is a variational autoencoder with a categorical distribution over the latent space. The model consists of two parts: the encoder $E_\theta$ and decoder $D_\theta$.
The encoder first reduces the dimension of the input image from $128 \times 128$ pixels to $8 \times 8$ tokens by several strided convolution layers. The convolutional part is followed by a quantization layer, which maps the resulting feature map to a discrete space. 
The quantization layer stores $n_{lat}$ embedding vectors of the same dimension as the feature vectors returned by the convolutional part of the encoder.
It encodes each point of the feature map by returning the index of the closest embedding vector. The output of the encoder at position $(i, j)$ for image $x$ is: 
\begin{equation}\label{eq:codebook}
\argmin_k \, \lVert (f^{(enc)}_\theta(x))_{i, j} - W^{(emb)}_k \rVert_2\enspace,
\end{equation}
where $W^{(emb)} \in \mathbb{R}^{n_{lat} \times d_{lat}}$ is the embedding matrix with rows $W_k^{(emb)}$ of length $d_{lat}$ and $f_\theta^{(enc)}$ is the convolutional part of the encoder. The decoder then performs an inverse operation by first encoding the indices back to the embedding vectors by using $W^{(emb)}$ followed by several convolutional layers combined with upscaling to increase the spatial dimension back to the original image size. 

Since the operation in \cref{eq:codebook} is not differentiable, we approximate the gradient with a straight-through estimator \cite{bengio2013estimating} and copy the gradients from the decoder input to the encoder output.
The final loss for the codebook is a weighted sum of three parts: the pixel-wise mean absolute error (MAE) between the input image and the reconstructed image, the perceptual loss between the input and reconstructed image \cite{esser2021taming}, and the commitment loss \cite{oord2017neural,razavi2019generating} $\mathcal{L}_c$, which encourages the output of the encoder to stay close
to the chosen embedding vector to prevent it from fluctuating too frequently from one vector
to another:
\begin{equation}
\mathcal{L}_c = \min_{k} \,|| f_{\theta}^{(enc)}(x)_{i, j} - \text{sg}(W_k^{(emb)})||_2^2 \enspace,
\end{equation}
where $\text{sg}$ is the stop-gradient operation \cite{oord2017neural}. We use the exponential moving average updates for the codebook \cite{oord2017neural}.
See \cite{oord2017neural,razavi2019generating} for more details on the codebook, and the \suppmat for the architecture details.

\PAR{Transformer.}\label{sec:transformer}%
We first describe the case of image generation and extend the approach to camera pose estimation later. We optimize the transformer for multiple context sizes and multiple query views in the batch at the same time. This has two benefits: it will allow the trained model to handle different context sizes, and the model will fully utilize the training batch (multiple images will be targets in the loss function).
Each training batch consists of a set of $n$ views. 
Let $(x_i)_{i=1}^{n}$ be the sequence of images under a random ordering and $(c_i)_{i=1}^{n}$ be the sequence of the associated camera poses.
Let us also define the sequence of images transformed by the encoder $E_\theta$ parametrized by $\theta$ as $s_i = E_\theta(x_i)$, $i=1, \ldots, n$.
Note that each $s_i$ is itself a sequence of tokens.
With this formulation, we generate the next image in the sequence given all the previous views, effectively optimizing all different context sizes at once. Therefore, we model the probability $p(s_i|s_{<i}, c_{\leq i})$.
Note that we do not optimize the first $n_{\text{min}}$ views (called the \emph{pure context}), because they usually do not provide enough information for the task. 

\begin{figure}[t]
\definecolor{loccolor}{HTML}{ff2128}
\definecolor{gencolor}{HTML}{0899c1}
\centering
\includegraphics[width=0.55\linewidth,trim={0 0.2cm 0 0.2cm},clip]{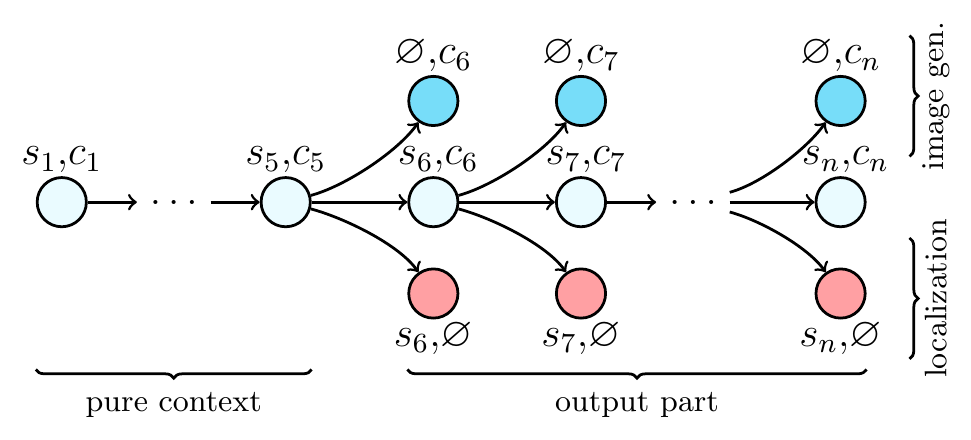}
\caption{
\textbf{Branching attention} mechanism: the nodes represent parts of the processed sequence. Starting in any node and tracing the arrows backwards gives the sequence over which the attention is computed, \eg, node $s_7,\varnothing$ attends to $s_1, c_1$, $s_2, c_2$, \ldots, $s_7, \varnothing$. \textcolor{gencolor}{Blue} and \textcolor{loccolor}{red} nodes in the last transformer block are used in the loss computation
\label{fig:branching-attention}}
\end{figure}

In practice, we need to replace the tokens corresponding to each query view with mask tokens to allow the transformer to decode them in a single forward pass. 
For the image generation task, the tokens of the last image in the sequence are replaced with special mask tokens $\lambda$, and, for the localization task, the tokens of the last image do not include the camera pose (denoted as $\varnothing$).
However, if we replaced the tokens in the training batch, the next query image would not be able to perceive the original tokens. Therefore, we have to process both the original and the masked tokens.
For the $i$-th query image, we need the sequence of $i - 1$ context views ending with masked tokens at the $i$-th position. We can represent the sequences as a tree (see \cref{fig:branching-attention}) where different endings branch off the shared trunk.
By following a leaf node back to the root of the tree, we recover the original sequence corresponding to the particular query view.

For localization, we train the model to output the camera pose $c_i$ given $s_{\leq i}$ and  $c_{<i}$.
For image generation, this leads to $n - n_{\text{min}}$ sequences. We attach a regression head to the hidden representation of all tokens of the last image in the sequence. 
The query image tokens form the input, and we mask the camera poses by replacing the camera pose representation with a single trainable vector.

\PAR{Branching attention.}
In this section, we introduce the \emph{branching attention} which computes attention over the tree shown in \cref{fig:branching-attention}, and allows us to optimize the transformer model for all context sizes and tasks very efficiently.
Note that we have to forward all tree nodes through all layers of the transformer. Therefore, the memory and time complexity is proportional to the number of nodes in the tree and thus to the number of views and tasks.

The input to the branching attention is a sequence of triplets of keys, values, and queries: $\big((K^{(i)}, Q^{(i)}, V^{(i)})\big)_{i=0}^{p}$ for $p = 2$, because we train the model on two tasks. Each element in the sequence corresponds to a single row in \cref{fig:branching-attention} and $i = 0$ is the middle row.
All $K^{(i)}$, $Q^{(i)}$, $V^{(i)}$ have the size $(n k^2) \times d_m$ where $d_m$ is the dimensionality of the model and $k$ is the size of the image in the latent space.
The output of the branching attention is a sequence $\big(R^{(i)}\big)_{i=0}^{p}$. 
The case of $R^{(0)}$ is handled differently because it corresponds to the trunk shared for all tasks and context sizes. 
Let us define a lower triangular matrix $M \in \mathbb{R}^{n \times n}$, where $m_{i,j} = 1$ if $i \le j$. We compute the causal block attention
as:
\begin{equation}
R^{(0)} = (\text{softmax}(Q^{(0)} (K^{(0)})^T) \odot M \otimes \mathbf{1}^{k^2 \times k^2}) V^{(0)}\enspace, \label{eqn:block_att}
\end{equation} 
where $\otimes$ and $\odot$ are the Kronecker and element-wise product, respectively, and $\mathbf{1}^{m \times n}$ is a matrix of ones. 
\cref{eqn:block_att} is similar to normal masked attention \cite{vaswani2017attention} with the only difference in the causal mask. In this case, we allow the model to attend to all previous images and all other vectors from the same image.
For $i > 0$ we can compute $R^{(i)}$ as follows:
\begin{align}
D &= Q^{(i)} (K^{(0)})^T\enspace,\\
C &= \begin{bmatrix} Q_{1:k^2}^{(i)} (K_{1:k^2}^{(i)})^T\\
\vdots \\
Q_{(n - 1) \cdot k^2 + 1:n \cdot k^2}^{(i)} (K_{(n - 1) \cdot k^2 + 1:n \cdot k^2}^{(i)})^T
\end{bmatrix}\enspace,\\
S &= \text{softmax}([D, C]) \odot [(M - I) \otimes \mathbf{1}^{k^2 \times k^2}), \mathbf{1}^{nk^2 \times k^2}]\enspace,\\
S' &= S_{\cdot, 1:n \cdot k^2}\,, S'' = S_{\cdot, n \cdot k^2 + 1: (n + 1) \cdot k^2}\enspace,\\
R^{(i)} &= S' V^{(0)} + \begin{bmatrix}
S''_{1:k^2} V^{(i)}_{1:k^2}\\
\vdots\\
S''_{n \cdot k^2 + 1: (n + 1) \cdot k^2} V^{(i)}_{n \cdot k^2 + 1:(n+1) \cdot k^2}\\
\end{bmatrix}\enspace.
\end{align}
Matrix $D$ represents the unmasked raw attention scores between $i$-th queries and keys from all previous images. Matrix $C$ contains the raw pairwise attention scores between $i$-th queries and $i$-th keys (the ending of each sequence). Then, the softmax is computed to normalize the attention scores and the causal mask is applied to the result, yielding the attention matrix $S$, and the respective values are weighted by the computed scores. In particular,
the scores contained in the last $k^2$ columns of the attention matrix
are redistributed back to the associated $i$-th values.
The result $R^{(0)}$ corresponds to the nodes in the middle row in \cref{fig:branching-attention}, whereas $R^{(i)}, i > 0$ are the other nodes.

\PAR{Transformer input and training.}
To build the input for the transformer, we first embed all image tokens into trainable vector embeddings of length $d_m$.
Before passing camera poses to the network, we express all camera poses relative to the first context camera pose in the sequence. We represent camera poses by concatenating the 3D position with the normalized orientation quaternion (a unit quaternion with a positive real part). 
Finally, we transform the camera poses with a trainable feed-forward neural network in order to increase the dimension to the same size as image token embeddings $d_m$ in order to be able to sum them. 

 Similarly to \cite{radford2019language}, we also add the positional embeddings by summing the input sequence with a sequence of trainable vectors. However, our positional embeddings are shared for all images in the sequence, \ie, the $i$-th token of every image will share the same positional embedding.

The output of the last transformer block is passed to an affine layer followed by a softmax layer, and it is trained using the cross-entropy loss to recover the last $k^2$ tokens ($s_{j,1}, \ldots, s_{j,k^2}$). For the localization task, the output is passed through a two-layer feed-forward neural network, and it is trained using the mean square error to match the ground-truth camera pose of the last $k^2$ tokens. Note that we compute the losses over position and orientation separately and add them together without weighing.\footnote{We tried dynamic weighting as described in \cite{kendall2017geometric}, but it performed worse.}
Since we attach the pose prediction head to the hidden representation of all tokens of the query image, we obtain multiple pose estimates. During inference, we simply average them. 

\section{Experiments}\label{sec:experiments}
To answer the question of whether explicit 3D reasoning is really needed for novel view synthesis, we designed a series of experiments evaluating the proposed approach.
First, we evaluate the codebook, whose performance is the upper bound on what we can achieve with the full pipeline. 
We next compare our method to GQN-based methods \cite{eslami2018neural,tobin2019geometry,chen2021direct} that also do not use continuous volumetric scene representations. We continue by evaluating our approach on other synthetic data. Then, we compare our approach to state-of-the-art NeRF-based approaches on a real-world dataset. Finally, we show our model's localization performance.

We evaluate our approach on both real and synthetic datasets:
a) \textbf{Shepard-Metzler-7-Parts (SM7)} \cite{shepard1971mental,eslami2018neural} is a synthetic dataset, where objects composed of 7 cubes of different colors are rotated in space.
b) \textbf{ShapeNet} \cite{chang2015shapenet} is a synthetic dataset of simple objects. We use $128 \times 128$ pixel images rendered by \cite{sitzmann2019scene} containing two categories: cars and chairs.
c) \textbf{InteriorNet} \cite{li2018interiornet} is a collection of interior environments designed by 1,100 professional designers. We used the publicly available part of the dataset 
(20k scenes with 20 images each).
While the dataset is synthetic, the renderings are 
similar to real-world environments.
The first 600 environments serve as our test set.
d) \textbf{Common Objects in 3D (CO3D)} \cite{reizenstein2021common} is a real-world dataset containing 1.5 million images showing almost 19k objects from 51 MS-COCO~\cite{lin2014microsoft} categories (\eg, apple, donut, vase, 
etc.). 
The capture of the dataset was crowd-sourced.
e) \textbf{7-Scenes} \cite{glocker2013real} is a real-world dataset depicting 7 indoor scenes as captured by a  Kinect RGB-D camera. The dataset consists of 44 sequences of 500--1,000 frames each and it is a standard benchmark for visual localization~\cite{balntas2018relocnet,laskar2017camera,melekhov2017relative,kendall2015posenet,brahmbhatt2018geometry}.

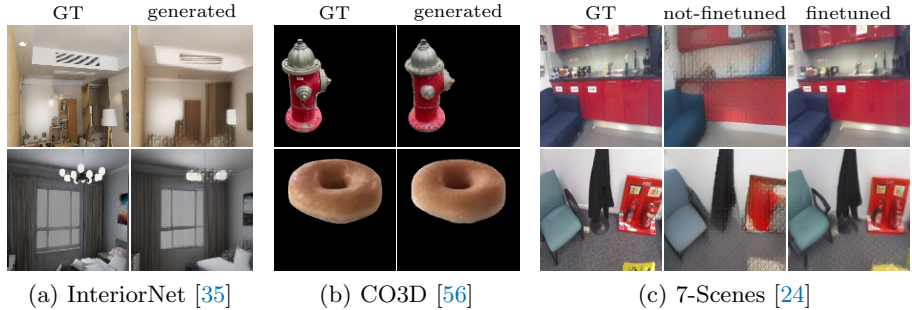
\begin{figure}[t!]
\newlength\imgwid
\setlength\imgwid{0.132\linewidth}
\centering
\subfloat[InteriorNet \cite{li2018interiornet}]{
    \centering
    \begin{tikzpicture}[
     image/.style = {text width=\imgwid, 
                     inner sep=0pt, outer sep=0pt},
    label/.style = { minimum height=0.4cm },
    node distance = 1pt and 1pt
                            ] 
    \scriptsize
    \path coordinate(last);
    \node [image,below=of last,alias=last] (img00)
    {\includegraphics[width=\linewidth]{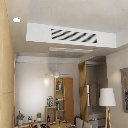}};
    \node [image,right=of img00] (img01)
    {\includegraphics[width=\linewidth]{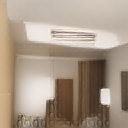}};
    
    \node [image,below=of last,alias=last] (img10)
    {\includegraphics[width=\linewidth]{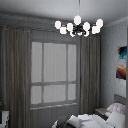}};
    \node [image,right=of img10] (img11)
    {\includegraphics[width=\linewidth]{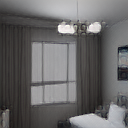}};
    
    \node [label,above=-1pt of img00] {GT};
    \node [label,above=-1pt of img01] {generated};
    \end{tikzpicture}
}
\hfill
\subfloat[CO3D \cite{reizenstein2021common}]{
    \centering
    \begin{tikzpicture}[
     image/.style = {text width=\imgwid, 
                     inner sep=0pt, outer sep=0pt},
    label/.style = { minimum height=0.4cm },
    node distance = 1pt and 1pt
                            ] 
    \scriptsize
    \path coordinate(last);
    \node [image,below=of last,alias=last] (img00)
{\includegraphics[width=\linewidth]{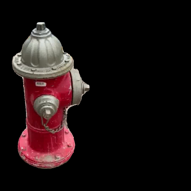}};
    \node [image,right=of img00] (img01)
{\includegraphics[width=\linewidth]{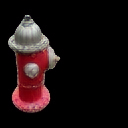}};
    
    \node [image,below=of last,alias=last] (img10)
{\includegraphics[width=\linewidth]{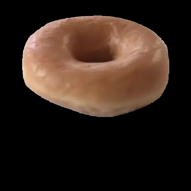}};
    \node [image,right=of img10] (img11)
{\includegraphics[width=\linewidth]{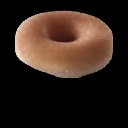}};
    \node [label,above=-1pt of img00] {GT};
    \node [label,above=-1pt of img01] {generated};
    \end{tikzpicture}
}
\hfill
\subfloat[7-Scenes \cite{glocker2013real}]{
    \centering
    \begin{tikzpicture}[
     image/.style = {text width=\imgwid, 
                     inner sep=0pt, outer sep=0pt},
    label/.style = { minimum height=0.4cm },
    node distance = 1pt and 1pt
                            ] 
    \scriptsize
    \path coordinate(last);
    \node [image,below=of last,alias=last] (img00)
    {\includegraphics[width=\linewidth,height=\linewidth]{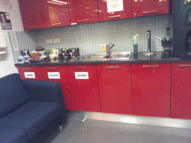}};
    \node [image,right=of img00] (img01)
    {\includegraphics[width=\linewidth]{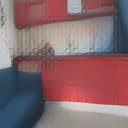}};
    \node [image,right=of img01] (img02) 
    {\includegraphics[width=\linewidth]{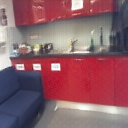}};

    \node [image,below=of last,alias=last] (img10)
    {\includegraphics[width=\linewidth,height=\linewidth]{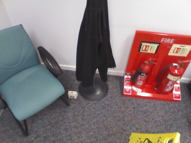}};
    \node [image,right=of img10] (img11)
    {\includegraphics[width=\linewidth]{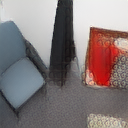}};
    \node [image,right=of img11] (img12) 
    {\includegraphics[width=\linewidth]{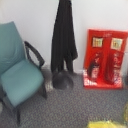}};
    
    \node [label,above=-1pt of img00] {GT};
    \node [label,above=-1pt of img01] {not-finetuned};
    \node [label,above=-1pt of img02] {finetuned};
    \end{tikzpicture}
}
\caption{\textbf{Codebook evaluation} on multiple datasets comparing the ground truth (\textbf{GT}) with the reconstructed image. For the 7-Scenes dataset, we compare the model finetuned and not-finetuned on the 7-Scenes dataset}
\label{fig:codebook}
\end{figure}
\begin{figure}[t]
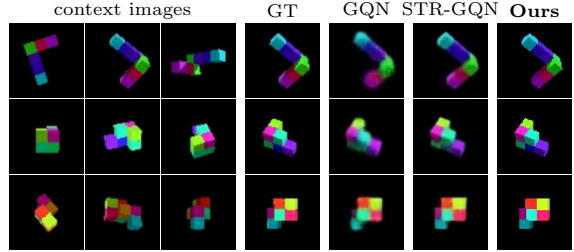

\scriptsize
\centering
\begin{tikzpicture}[
 image/.style = {text width=0.08\linewidth, 
                 inner sep=0pt, outer sep=0pt},
label/.style = { minimum height=0.4cm },
node distance = 1pt and 1pt
                        ] 
\path coordinate(last);
\foreach \i in {0,1,...,2} {
    \node [image,below=of last,alias=last] (img\i0)
    {\includegraphics[width=\linewidth]{resources/gqn/c\i_0.png}};
\node [image,right=of img\i0] (img\i1)
    {\includegraphics[width=\linewidth]{resources/gqn/c\i_1.png}};
\node [image,right=of img\i1] (img\i2) 
    {\includegraphics[width=\linewidth]{resources/gqn/c\i_2.png}};
    \node [image,right=4pt of img\i2] (img\i3) 
    {\includegraphics[width=\linewidth]{resources/gqn/gt\i.png}};
    \node [image,right=4pt of img\i3] (img\i4) 
    {\includegraphics[width=\linewidth]{resources/gqn/gqn\i.png}};
    \node [image,right=4pt of img\i4] (img\i5) 
    {\includegraphics[width=\linewidth]{resources/gqn/str-gqn\i.png}};
    \node [image,right=4pt of img\i5] (img\i6)
    {\includegraphics[width=\linewidth]{resources/gqn/gen\i.png}};
}
\node [label,above=-1pt of img01] {context images};
\node [label,above=-1pt of img03] {GT};
\node [label,above=-1pt of img04] {GQN};
\node [label,above=-1pt of img05] {STR-GQN};
\node [label,above=-1pt of img06] {\textbf{Ours}};
\end{tikzpicture}
\caption{Results on the SM7 dataset. We compare against GQN \cite{eslami2018neural} and STR-GQN \cite{chen2021str}
\label{fig:migt-sm7}}
\end{figure}
\PAR{Codebook evaluation.}
First, we evaluate the quality of our codebooks by measuring the quality of the images generated by the encoder-decoder architecture without the transformer.
We trained codebooks of size 1,024 using the same hyperparameters for all experiments using an architecture very similar to~\cite{esser2021taming}.
The training took roughly 480~GPU-hours. A detailed description of the model and the hyperparameters is given in \suppmat as well as in the published code.

Examples of reconstructed images are shown in \cref{fig:codebook}.
As can be seen, although losing some details and image sharpness, the codebooks can recover the overall shape well. The results show that using the codebook leads to good results, even though we use only $8\times8$ codes to represent an image.
In some images, there are noticeable artifacts. In our analysis, we pinpointed the perceptual loss to be the cause, but removing the perceptual loss led to more blurry images.
Further analysis of the codebooks is included in the \suppmat

\PAR{Full method evaluation.}
The transformer is trained using only the tokens generated by the codebook.
Having verified that our codebooks work as intended, we evaluate our complete approach in the context of image synthesis.
The architecture of our transformer model is based on GPT2 \cite{radford2019language}. We give more details on the architecture, the motivation, and the hyperparameters in the \suppmat

\noindent The \textbf{SM7} dataset was used to compare our approach to other methods that only operate in 2D image space~\cite{eslami2018neural,tobin2019geometry,chen2021str}.
Our method achieved the best mean absolute error (MAE) of \textbf{1.61}, followed by E-GQN \cite{tobin2019geometry} with 2.14, STR-GQN \cite{chen2021direct} with 3.11 and the original GQN \cite{eslami2018neural} method with MAE 3.13. The results were averaged over 1,000 scenes (context size was 3) and computed on images with size $64 \times 64$ pixels. A qualitative comparison is shown in \cref{fig:migt-sm7}.

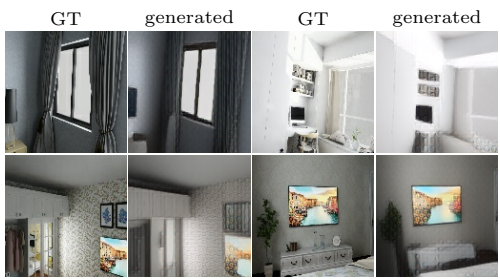
\begin{figure}[t]
\centering
\begin{tikzpicture}[
 image/.style = {text width=0.132\linewidth, 
                 inner sep=0pt, outer sep=0pt},
label/.style = { minimum height=0.4cm },
node distance = 1pt and 1pt
                        ] 
\scriptsize
\path coordinate(last);
\node [image,below=of last,alias=last] (img00)
{\includegraphics[width=\linewidth]{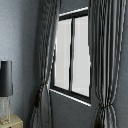}};
\node [image,right=of img00] (img01)
{\includegraphics[width=\linewidth]{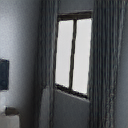}};
\node [image,right=of img01] (img02) 
{\includegraphics[width=\linewidth]{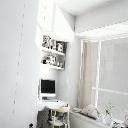}};
\node [image,right=of img02] (img03) 
{\includegraphics[width=\linewidth]{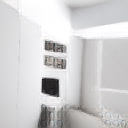}};

\node [image,below=of last,alias=last] (img10)
{\includegraphics[width=\linewidth]{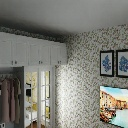}};
\node [image,right=of img10] (img11)
{\includegraphics[width=\linewidth]{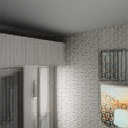}};
\node [image,right=of img11] (img12) 
{\includegraphics[width=\linewidth]{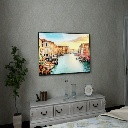}};
\node [image,right=of img12] (img13) 
{\includegraphics[width=\linewidth]{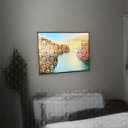}};
\node [label,above=-1pt of img00] {GT};
\node [label,above=-1pt of img01] {generated};
\node [label,above=-1pt of img02] {GT};
\node [label,above=-1pt of img03] {generated};
\end{tikzpicture}
\caption{Evaluation of our method on the InteriorNet dataset with the context size 19
\label{fig:interiornet}}
\end{figure}
\noindent We use the \textbf{InteriorNet} dataset because of its large size and realistic appearance. 
The models pre-trained on it are also used in other experiments. Since each scene provides 20 images, we use 19 context views.
\cref{fig:interiornet} shows images generated by the model trained for both the localization and novel view synthesis tasks.

\begin{figure}[t]
    \centering
    \newcommand\figrowwide[2]{
    \node [image,below=of last,alias=last] (img0)
    {\includegraphics[width=\linewidth,trim={0 5mm 0 7mm},clip]{#1/gt/00-064.png}};
    \node [image,right=of img0] (img1)
    {\includegraphics[width=\linewidth,trim={0 5mm 0 7mm},clip]{#1/gt/01-104.png}};
    \node [image,right=of img1] (img2)
    {\includegraphics[width=\linewidth,trim={0 5mm 0 7mm},clip]{#1/gt/#2.png}};
    \node [image,right=of img2] (img3) 
    {\includegraphics[width=\linewidth,trim={0 5mm 0 7mm},clip]{#1/pixelnerf/000#2.png}};
    \node [image,right=0.3cm of img3] (img4) 
    {\includegraphics[width=\linewidth,trim={0 5mm 0 7mm},clip]{#1/viewformer/#2.png}};
}
\begin{tikzpicture}[
     image/.style = {text width=0.12\textwidth, 
                     inner sep=0pt, outer sep=0pt},
    label/.style = {  },
    node distance = 0pt and 1pt
                            ]

    \scriptsize
    \path coordinate(last);
    
    \figrowwide{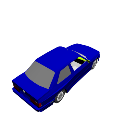}{000}
    \node [label,above=-5.5pt of $(img0.north)!0.5!(img1.north)$] {context views};
    \node [label,above=-6pt of img2] {GT};
    \node [label,above=-6pt of img3] {PixelNeRF};
    \node [label,above=-6pt of img4] {\textbf{ViewFormer}};
    \figrowwide{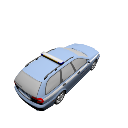}{025}
    \figrowwide{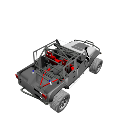}{030}
    \figrowwide{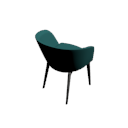}{164}
    \figrowwide{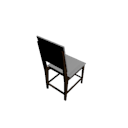}{040}
    \figrowwide{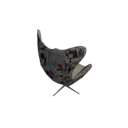}{000}
    
    \draw[dashed] ([yshift=5.15cm]img1.north east) -- ([yshift=0cm]img1.south east); 
    \end{tikzpicture}

    \caption{
    \textbf{ShapeNet} qualitative comparison with PixelNeRF
    \cite{yu2021pixelnerf} using 2 context views
\label{fig:shapenet}
}
\end{figure}
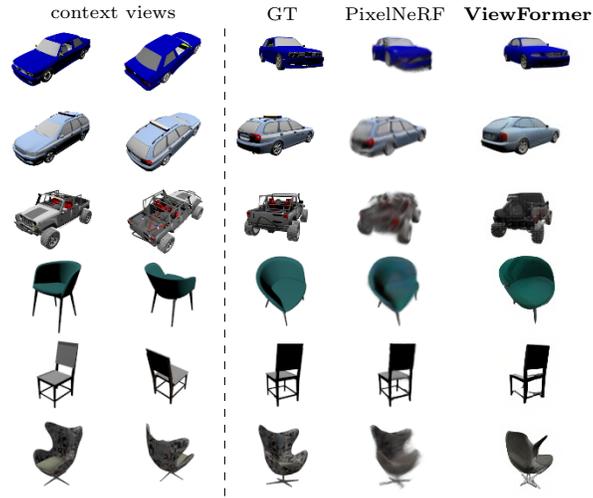

\PAR{ShapeNet evaluation.}We used the InteriorNet pre-trained model and we fine-tuned it on the ShapeNet dataset. We trained a single model for both categories (cars and chairs) using 3 context views. The training details and additional results are given in \suppmat We show the qualitative comparison with PixelNeRF \cite{yu2021pixelnerf} in \cref{fig:shapenet}. PixelNeRF trained a different model for each category.

The results show that our method achieves good visual quality overall, especially on the cars dataset. However, the geometry is slightly distorted on the chairs. Compared to PixelNeRF, it prefers to hallucinate a part of the scene instead of rendering a blurry image.
This can cause some neighboring views to have a different color or shape in places where the scene is less covered by context views. 
However, this problem can be reduced by simply adding the previously generated view to the set of context views.
See the video in the \suppmat

\begin{table*}[t!]
\caption{
\textbf{Novel view synthesis} results on the CO3D dataset \cite{reizenstein2021common} on all categories and 10 categories from \cite{reizenstein2021common}. We compare ViewFormer with and without localization (`no-loc') trained on all categories (`@ all cat.') and 10 selected categories (`@ 10 cat.'). We show the PSNR and LPIPS for seen and unseen scenes (`train' and `test') and
test PSNR with varying context size. The best value is \textbf{bold}; the second is \underline{underlined}
\label{tab:co3d_all}}
\newcommand{\setfont}{}%
\newcommand{\testset}{{\setfont test}\xspace}
\newcommand{\trainset}{{\setfont train}\xspace}
\newcommand{\traintrainset}{{\setfont train-known}\xspace}
\newcommand{\traintestset}{{\setfont train}\xspace}
\newcommand{\testtrainset}{{\setfont test-known}\xspace}
\newcommand{\testtestset}{{\setfont test}\xspace}

\newcommand{\method}[1]{{\oneptsmaller{\sffamily #1}}}
\newcommand{\methodrbtl}[1]{{\small{\sffamily #1}}}

\newcommand{\bb}[1]{{\textbf{#1}}}
\renewcommand{\b}[1]{{\underline{#1}}}
\newcommand{\transformerAnerfAopacityAdimdown}{\method{NerFormer}\cite{reizenstein2021common}}
\newcommand{\nerfAopacity}{\method{NeRF\cite{mildenhall2020nerf}} }
\newcommand{\nerfAautodecAopacity}{\method{NeRF+AD} }
\newcommand{\nerfAviewpoolAopacity}{\method{NeRF+WCE}\cite{henzler2021unsupervised}}
\newcommand{\nvAautodec}{\method{NV\cite{lombardi2019neural}}}
\newcommand{\nvAviewpool}{\method{NV+WCE}}
\newcommand{\idr}{\method{IDR\cite{yariv2020multiview}}}
\newcommand{\idrAautodec}{\method{IDR+AD}}
\newcommand{\srnAharmonicAautodecAopacity}{\method{SRN+$\gamma$}}
\newcommand{\srnAautodecAopacity}{\method{SRN\cite{sitzmann19scene}}}
\newcommand{\srnAharmonicAviewpoolAopacity}{\method{SRN+WCE+$\gamma$}}
\newcommand{\srnAviewpoolAopacity}{\method{SRN+WCE}}
\newcommand{\dvrAautodec}{\method{DVR\cite{niemeyer2020differentiable}}}
\newcommand{\dvrAharmonicAautodec}{\method{DVR+$\gamma$}}
\newcommand{\softrasAautodec}{\method{P3DMesh\cite{ravi2020pytorch3d}}}
\newcommand{\softrasAviewpool}{\method{P3DMesh}}
\newcommand{\pointcloudsAautodecAopacities}{\method{IPC}}
\newcommand{\pointcloudsAviewpoolAopacities}{\method{IPC+WCE}}

\setlength\tabcolsep{0.05cm}
\scriptsize\centering
\newcommand{\isad}{\textsuperscript{\textdagger}}
\begin{tabular}{
@{}
c@{\extracolsep{0.1cm}}
lr@{\extracolsep{0.1cm}}
cc@{\extracolsep{0.1cm}}
cc@{\extracolsep{0.1cm}}
ccccc@{\extracolsep{0.1cm}}
ccc
@{}
}%

               & & & \multicolumn{2}{c}{avg. test} 
               & \multicolumn{2}{c}{avg. train} 
               & \multicolumn{5}{c}{PSNR$\uparrow$ @ \# ctx. size}\\
               \cmidrule{4-7} \cmidrule{8-12}

                EC & Method                        & 3D & PSNR$\uparrow$ & LPIPS$\downarrow$ & PSNR$\uparrow$ & LPIPS$\downarrow$ & 9  & 7 & 5 & 3  & 1 & \\ \cmidrule{1-3} \cmidrule{4-5} \cmidrule{6-7} \cmidrule{8-12}
                
\multirow{6}{*}{\STAB{\rotatebox[origin=c]{90}{all categories}}}
 & \textbf{\method{\Ours @ all cat.}} & \xmark  & 15.3   & \bb{0.23}   & 15.6   & \bb{0.22}   & 16.1   & 15.9   & 15.5   & 15.1   & 13.7   \\
 & \textbf{\method{\Ours no-loc @ all cat.}} & \xmark & \b{15.4}   & \bb{0.23}   & 15.8   & \bb{0.22}   & \b{16.2}   & \b{16.0}   & \b{15.6}   & \b{15.2}   & \b{13.8}   \\
 \cmidrule{2-3} \cmidrule{4-5} \cmidrule{6-7} \cmidrule{8-12}
 
 & \transformerAnerfAopacityAdimdown  & \xmark & \bb{15.7} & \b{0.24} & \b{16.5}  & \b{0.24} & \bb{16.7} & \bb{16.4} & \bb{16.1} & \bb{15.5} & \bb{13.9} \\
 & \srnAviewpoolAopacity             & \xmark & 14.2  & 0.27  & 16.3      & 0.25  & 14.4  & 14.3  & 14.3  & 14.2  & 13.5  \\
 & \srnAharmonicAviewpoolAopacity    & \xmark & 13.7      & 0.28      & \bb{17.1} & 0.25      & 14.0      & 13.8      & 13.9      & 13.7      & 13.2      \\
 & \nerfAviewpoolAopacity            & \xmark & 11.6      & 0.27  & 12.6      & 0.27      & 11.9      & 11.8      & 11.8      & 11.6      & 10.8      \\
\cmidrule{1-12}
\morecmidrules
\cmidrule{1-12}
\multirow{11}{*}{\STAB{\rotatebox[origin=c]{90}{10 categories}}}
 
& \textbf{\method{\Ours @ 10 cat.}} & \xmark  & 15.6   & \bb{0.25}   & 16.6   & \b{0.23}   & 16.5   & 16.3   & 15.8   & 15.3   & 14.0   \\
& \textbf{\method{\Ours no-loc @ 10 cat.}} & \xmark  & 15.6   & \bb{0.25}   & 17.1   & \bb{0.22}   & 16.5   & 16.2   & 15.8  & 15.3   & 14.0   \\

& \textbf{\method{\Ours @ all cat.}} & \xmark  & 16.0   & \bb{0.25}   & 16.4   & 0.24   & \b{17.0}   & 16.7   & \b{16.3}   & 15.7   & \b{14.3}   \\
& \textbf{\method{\Ours no-loc @ all cat.}} & \xmark & \b{16.1}   & \bb{0.25}   & 16.6   & \b{0.23}   & \b{17.0}   & \b{16.8}   & \b{16.3}   & \b{15.8}   & \b{14.3}   \\
 \cmidrule{2-3} \cmidrule{4-5} \cmidrule{6-7} \cmidrule{8-12}

& \transformerAnerfAopacityAdimdown    & \cmark            & \bb{17.6} & 0.27  & \bb{17.9} & 0.26   & \bb{18.9} & \bb{18.6} & \bb{18.1} & \bb{17.1} & \bb{15.1} \\
& \srnAharmonicAviewpoolAopacity      & \cmark            & 14.4      & 0.27  & \b{17.6}  & 0.24  & 14.6      & 14.5      & 14.6      & 14.5      & 13.9  \\
& \srnAviewpoolAopacity                & \cmark           & 14.6  & 0.27 & 16.6      & 0.26     & 14.9  & 14.8  & 14.8  & 14.6      & 13.9  \\
& \nerfAviewpoolAopacity               & \cmark           & 13.8      & 0.27 & 14.3      & 0.27     & 12.6      & 14.5      & 14.4      & 14.2      & 13.8      \\
& \pointcloudsAviewpoolAopacities        & \cmark         & 13.5      & 0.37     & 14.1      & 0.36     & 13.8      & 13.8      & 13.7      & 13.6      & 12.6      \\
& \softrasAviewpool                    & \cmark           & 12.4      & \b{0.26}& 17.2      & \b{0.23}& 12.6      & 12.5      & 12.5      & 12.5      & 12.1      \\
& \nvAviewpool                         & \cmark           & 11.6      & 0.35     & 12.3      & 0.34     & 11.7      & 11.6      & 11.6      & 11.6      & 11.3      \\\bottomrule
 
\end{tabular}%

\end{table*}

\PAR{Common Objects in 3D.}
In order to show that we can transfer a model pre-trained on synthetic data to real-world scenes, we evaluate our method on the CO3D dataset \cite{reizenstein2021common}. We compare our approach with NeRF-based methods using the results reported in~\cite{reizenstein2021common}.
Unfortunately, we tried to train the PixelNeRF \cite{yu2021pixelnerf} on the CO3D dataset, but were not able to obtain good results. Therefore we omit it from the comparison.
While the baselines are trained separately per category, we train two transformer models: 
one on the 10 categories used for evaluation in \cite{reizenstein2021common} and one for all dataset categories.
We fine-tune the model trained on the InteriorNet dataset. The context size is 9. Additional details and hyperparameters are given in \suppmat

The testing set of each category in the CO3D dataset is split into two subsets: `train' and `test' containing unseen images of objects seen and unseen during training respectively.
We use the evaluation procedure provided by Reizenstein \etal \cite{reizenstein2021common}. It evaluates the model on 1,000 sequences from each category with context sizes 1, 3, 5, 7, 9. The PSNR) and the LPIPS distance \cite{zhang2018unreasonable} are reported. Note that the PSNR is calculated only on foreground pixels. For more details on the evaluation procedure and the details of compared methods, please see~\cite{reizenstein2021common}.

\cref{tab:co3d_all} shows results of the evaluation on all CO3D categories and on the 10 categories used for evaluation in 
\cite{reizenstein2021common}.
Our method is competitive even though it does not explicitly reason in 3D as other baselines, does not utilize object masks, and even though we trained a single model for all categories while other baselines are trained per category. 
Note that on the whole dataset, the top-performing method, NerFormer~\cite{reizenstein2021common}, was trained for about 8400~GPU-hours while training our codebook 
took 480~GPU-hours, training the transformer on InteriorNet took 280~GPU-hours, and fine-tuning the transformer took 90~GPU-hours, giving a total of 850~GPU-hours.
Also, note that rendering a single view takes 178\,s for the NerFormer and only 93\,ms for our approach.

The results show that our model has a large capacity (it is able to learn all categories while the baselines are only trained on a single category), and it benefits from more training data as can be seen when comparing models trained on 10 and all categories. We also observe that models achieve a higher performance on 10 categories than on all categories, suggesting that the categories selected by the authors of the dataset are easier to learn or of higher quality.
All our models outperform all baselines in terms of LPIPS, which indicates that the images can look more realistic while possibly not matching the real images very precisely.

\begin{figure}[t!]
\centering
\begin{tikzpicture}[
 image/.style = {text width=0.132\linewidth, 
                 inner sep=0pt, outer sep=0pt},
label/.style = { minimum height=0.4cm },
node distance = 1pt and 1pt
                        ] 
\scriptsize
\path coordinate(last);
\node [image,below=of last,alias=last] (img00)
{\includegraphics[width=\linewidth]{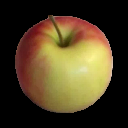}};
\node [image,right=of img00] (img01)
{\includegraphics[width=\linewidth]{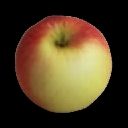}};
\node [image,right=of img01] (img02) 
{\includegraphics[width=\linewidth]{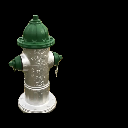}};
\node [image,right=of img02] (img03) 
{\includegraphics[width=\linewidth]{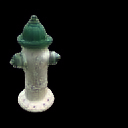}};

\node [image,below=of last,alias=last] (img10)
{\includegraphics[width=\linewidth]{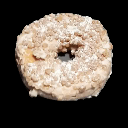}};
\node [image,right=of img10] (img11)
{\includegraphics[width=\linewidth]{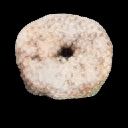}};
\node [image,right=of img11] (img12) 
{\includegraphics[width=\linewidth]{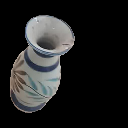}};
\node [image,right=of img12] (img13) 
{\includegraphics[width=\linewidth]{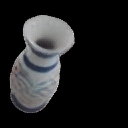}};
\node [label,above=-1pt of img00] {GT};
\node [label,above=-1pt of img01] {generated};
\node [label,above=-1pt of img02] {GT};
\node [label,above=-1pt of img03] {generated};
\end{tikzpicture}
\caption{Evaluation of our method on the CO3D dataset \cite{reizenstein2021common} with the context size 9}
\label{fig:co3d}
\end{figure}
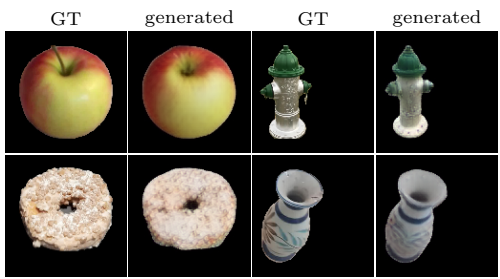

Fig.~\ref{fig:teaser} and~\ref{fig:co3d} show qualitative results. 
Our method is able to generalize well to unseen object instances, although it tends to lose some details. 
To answer the original question if explicit 3D reasoning is needed for novel view synthesis, based on our results, we claim that even without explicit 3D reasoning, we can achieve similar results, especially when the data are noisy, \eg a real-world dataset.

\PAR{Evaluating localization accuracy on 7-Scenes.}
We compare the localization part of our approach to methods from the literature on the 7-Scenes dataset~\cite{glocker2013real}. 
Due to space constraints, here we only summarize the results of the comparisons. Detailed results can be found in the \suppmat

Our approach performs similar to existing APR and RPR techniques that also use only a single forward pass in a network~\cite{brahmbhatt2018geometry,kendall2015posenet,balntas2018relocnet,shavit2021learning}, but worse than iterative approaches such as~\cite{ding2019camnet} or methods that use more densely spaced synthetic views as additional input~\cite{moreau2021lens}. 
Note that these approaches that do not use 3D scene geometry are less accurate than state-of-the-art methods based on 2D-3D correspondences~\cite{sattler2016efficient,brachmann2020ARXIV,sarlin2019cvpr}. 
Overall, the results show that our approach achieves a similar level of pose accuracy as comparable methods.
Furthermore, our approach is able to perform both localization and novel view synthesis in a simple forward pass, while other methods can only be used for localization.

\section{Conclusions \& future work}
This paper presents a two-stage approach to novel view synthesis from a few sparsely distributed context images.
We train our model on classes of similar 3D scenes to be able to generalize to a novel scene with only a handful of images as opposed to NeRF and similar methods that are trained per scene.
The model consists of a VQ-VAE codebook \cite{oord2017neural} and a transformer model.
To efficiently train the transformer,
we propose a novel branching attention module.
Our approach, ViewFormer, 
can render a view from a previously unseen scene in 93\,ms
without any explicit 3D reasoning
and we train a single model to render multiple categories of objects, whereas NeRF-based approaches train per-category models \cite{reizenstein2021common}.
We show that our method is competitive with SoTA NeRF-based approaches especially on real-world data, even without any explicit 3D reasoning.
This is an intriguing result because it implies that either current NeRF-based methods are not utilizing the 3D priors effectively or that a 2D-only model is able to learn it on its own without explicit 3D modeling.
The experiments also show that ViewFormer outperforms other 2D-only multi-view methods.

One limitation of our approach is the large amount of data needed, which we tackle through pre-training on a large synthetic dataset. 
Also, we need to fine-tune both the codebook and the transformer to achieve high-quality results on new datasets, which could be resolved by utilizing a larger codebook trained on more data.
Using more tokens to represent images should increase the rendering quality and pose accuracy. 
We also want to experiment with a simpler architecture with no codebook and larger scenes, possibly of outdoor environments.

\vspace{0.8\baselineskip}
\noindent\textbf{Acknowledgements.}
This work was supported by the European Regional Development Fund under projects 
IMPACT (reg. no. CZ.02.1.01/0.0/0.0/15\_003/ 0000468) and 
Robotics for Industry 4.0 (reg. no.  CZ.02.1.01/0.0/0.0/15\_003/ 0000470), 
the EU Horizon 2020 project RICAIP (grant agreement No 857306), the Grant Agency of the Czech Technical University in Prague (grant no. SGS22/ 112/OHK3/2T/13), and the Ministry of Education, Youth and Sports of the Czech Republic through the e-INFRA CZ (ID:90140).

\makeatletter
\@ifundefined{arxivPaper}
  {%
\clearpage                                         
\bibliographystyle{splncs04}
\bibliography{bibliography}
\end{document}
  }
  {}
\makeatother

\clearpage
\bibliographystyle{splncs04}
\bibliography{bibliography}
\clearpage

\renewcommand\thesection{\Alph{section}}
\renewcommand\thesubsection{\thesection.\arabic{subsection}}

\title{ViewFormer: NeRF-free Neural Rendering\\ from Few Images
Using Transformers\\ --\\ Supplementary Material}

\titlerunning{ViewFormer -- Supplementary Material}
\author{Jonáš Kulhánek \inst{1,2}\orcidlink{0000-0002-8437-3626} \and
Erik Derner\inst{1}\orcidlink{0000-0002-7588-7668} \and
Torsten Sattler\inst{1}\orcidlink{0000-0001-9760-4553} \and
Robert Babuška\inst{1,3}\orcidlink{0000-0001-9578-8598}}
\authorrunning{J. Kulhánek et al.}
\institute{Czech Institute of Informatics, Robotics and Cybernetics,\\
Czech Technical University in Prague
\and
Faculty of Electrical Engineering, Czech Technical University in Prague \and
Cognitive Robotics, Faculty of 3mE, Delft University of Technology \\
\vspace{0.4\baselineskip}
\url{https://jkulhanek.github.io/viewformer}
}

\maketitle

\setcounter{figure}{8}    
\setcounter{table}{1}    

\noindent In this supplementary material, we give more details on the results presented in the main paper and provide more details on the network architecture.
First, in \cref{sec:qualitative_results}, we present additional qualitative results on various datasets. We also show examples of context views used to render the final view.
The attached video is described in \cref{sec:attached_video}.
We include the camera pose estimation results on the 7-Scenes dataset \cite{glocker2013real} in \cref{sec:7scenes}, and we also show qualitative results of the novel view synthesis task on the same dataset.
In \cref{sec:ablation_study}, we present an ablation study. We also show how the performance increases with larger context sizes.
In Sections~\ref{sec:shapenet} and \ref{sec:sm7_evaluation}, we include additional results on the ShapeNet dataset and the Shepard-Metzler-Parts-7 (SM7) dataset, respectively.
Quantitative results of the codebook model are given in \cref{sec:codebook_evaluation}.
Finally, we give details on the training hyperparameters and architecture of the models in Sections~\ref{sec:training_details} and~\ref{sec:codebook_architecture}.

\section{Qualitative results}\label{sec:qualitative_results}
\begin{figure}[htbp]
\centering
\newcommand{\sz}{0.24\linewidth}
\begin{tikzpicture}[
 image/.style = {text width=2cm, anchor=north west,
                 inner sep=0pt, outer sep=0pt},
 simage/.style = {text width=1cm-0.5pt, anchor=north west,
                 inner sep=0pt, outer sep=0pt},
label/.style = { minimum height=0.4cm },
node distance = 1pt and 1pt
                        ] 
\scriptsize
\begin{scope}[]
\environmentTikzGroupEight{resources/supp/in-seq1/00000010}
\end{scope}
\begin{scope}[yshift=-2cm-1pt-1mm]
\environmentTikzGroupEight{resources/supp/in-seq2/00000051}
\end{scope}
\begin{scope}[yshift=-4cm-2pt-2mm]
\environmentTikzGroupEight{resources/supp/in-seq3/00000030}
\end{scope}

\node[label] at (0, 5pt) { context images };
\node [label] at (3cm+1pt, 5pt) {GT};
\node [label] at (5cm+1pt, 5pt) {generated};
\end{tikzpicture}
\caption{Visualization of the model trained on the InteriorNet dataset \cite{li2018interiornet}. We show the images generated with context size 8 while the model was trained with context size 19
\label{fig:apx_interiornet}}
\end{figure}
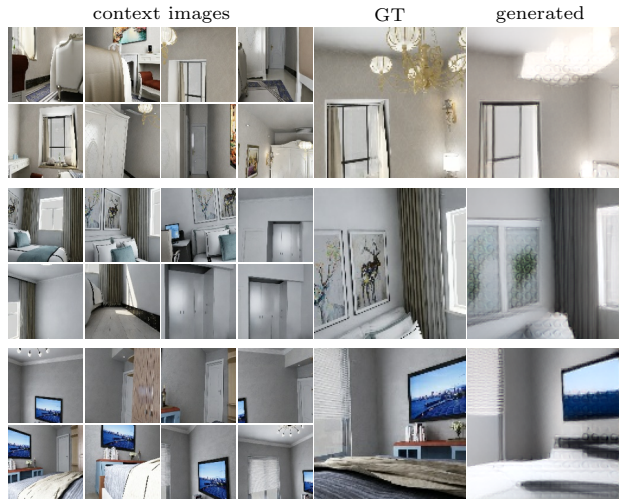

\begin{figure}[!t]
\centering
\newcommand{\sz}{0.24\linewidth}
\newcommand\figScale{0.85}
\newcommand\environmentTikzGroupEightCD[1]{
    \node [simage] at (0, 0)
    {\includegraphics[width=\figScale\linewidth]{#1-ctx/00.png}};
    \node [simage] at (1cm+0.5pt, 0)
    {\includegraphics[width=\figScale\linewidth]{#1-ctx/01.png}};
    \node [simage] at (0, -1cm-0.5pt)
    {\includegraphics[width=\figScale\linewidth]{#1-ctx/02.png}};
    \node [simage] at (1cm+0.5pt, -1cm-0.5pt)
    {\includegraphics[width=\figScale\linewidth]{#1-ctx/03.png}};
    \node [simage] at (-2cm-1.0pt, 0)
    {\includegraphics[width=\figScale\linewidth]{#1-ctx/04.png}};
    \node [simage] at (-1cm-0.5pt, 0)
    {\includegraphics[width=\figScale\linewidth]{#1-ctx/05.png}};
    \node [simage] at (-2cm-1.0pt, -1cm-0.5pt)
    {\includegraphics[width=\figScale\linewidth]{#1-ctx/06.png}};
    \node [simage] at (-1cm-0.5pt, -1cm-0.5pt)
    {\includegraphics[width=\figScale\linewidth]{#1-ctx/07.png}};
    \node [image] (gt) at (2cm+1pt, 0)
    {\includegraphics[width=\figScale\linewidth] {#1-gt.png}};
    \node [image] (gen) at (4cm+2pt, 0)
    {\includegraphics[width=\figScale\linewidth]{#1-gen-08.png}};
}

\newcommand\environmentTikzGroupFourCD[1]{
    \node [simage] at (0, 0)
    {\includegraphics[width=\figScale\linewidth]{#1-ctx/00.png}};
    \node [simage] at (1cm+0.5pt, 0)
    {\includegraphics[width=\figScale\linewidth]{#1-ctx/01.png}};
    \node [simage] at (0, -1cm-0.5pt)
    {\includegraphics[width=\figScale\linewidth]{#1-ctx/02.png}};
    \node [simage] at (1cm+0.5pt, -1cm-0.5pt)
    {\includegraphics[width=\figScale\linewidth]{#1-ctx/03.png}};
    \node [image] at (2cm+1pt, 0)
    {\includegraphics[width=\figScale\linewidth] {#1-gt.png}};
    \node [image] at (4cm+2pt, 0)
    {\includegraphics[width=\figScale\linewidth]{#1-gen-04.png}};
}

\newcommand\environmentTikzGroupTwoCD[1]{
    \node [image] at (0, 0)
    {\includegraphics[width=\figScale\linewidth]{#1-ctx/00.png}};
    \node [image] at (-2cm-1.0pt, 0)
    {\includegraphics[width=\figScale\linewidth]{#1-ctx/01.png}};
    \node [image] at (2cm+1pt, 0)
    {\includegraphics[width=\figScale\linewidth] {#1-gt.png}};
    \node [image] at (4cm+2pt, 0)
    {\includegraphics[width=\figScale\linewidth]{#1-gen-04.png}};
}

\newcommand\environmentTikzGroupOneCD[1]{
    \node [image] at (0, 0)
    {\includegraphics[width=\figScale\linewidth]{#1-ctx/00.png}};
    \node [image] at (2cm+1pt, 0)
    {\includegraphics[width=\figScale\linewidth] {#1-gt.png}};
    \node [image] at (4cm+2pt, 0)
    {\includegraphics[width=\figScale\linewidth]{#1-gen-04.png}};
}

\begin{tikzpicture}[
scale=\figScale,
 image/.style = {text width=2cm, anchor=north west,
                 inner sep=0pt, outer sep=0pt},
 simage/.style = {text width=1cm-0.5pt, anchor=north west,
                 inner sep=0pt, outer sep=0pt},
label/.style = { minimum height=0.4cm },
node distance = 1pt and 1pt
                        ] 
\scriptsize
\begin{scope}[]
\environmentTikzGroupOneCD{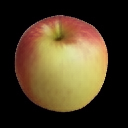}
\end{scope}
\begin{scope}[yshift=-2cm-1pt-1mm]
\environmentTikzGroupFourCD{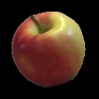}
\end{scope}
\begin{scope}[yshift=-4cm-2pt-2mm]
\environmentTikzGroupEightCD{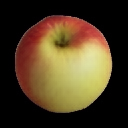}
\end{scope}
\begin{scope}[yshift=-6cm-3pt-3mm]
\environmentTikzGroupOneCD{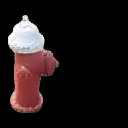}
\end{scope}
\begin{scope}[yshift=-8cm-4pt-4mm]
\environmentTikzGroupFourCD{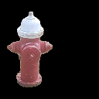}
\end{scope}
\begin{scope}[yshift=-10cm-5pt-5mm]
\environmentTikzGroupEightCD{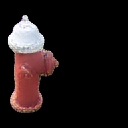}
\end{scope}
\begin{scope}[yshift=-12cm-6pt-6mm]
\environmentTikzGroupOneCD{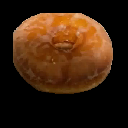}
\end{scope}
\begin{scope}[yshift=-14cm-7pt-7mm]
\environmentTikzGroupFourCD{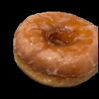}
\end{scope}
\begin{scope}[yshift=-16cm-8pt-8mm]
\environmentTikzGroupEightCD{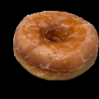}
\end{scope}

\node[label] at (1cm, 5pt) { ctx. images };
\node [label] at (3cm+1pt, 5pt) {GT};
\node [label] at (5cm+1pt, 5pt) {generated};
\end{tikzpicture}

\caption{Visualization of the model trained on the CO3D dataset \cite{reizenstein2021common}. We show the images generated with context sizes 1, 4, and 8 while the model was trained with context size 9
\label{fig:apx_co3d}}
\end{figure}
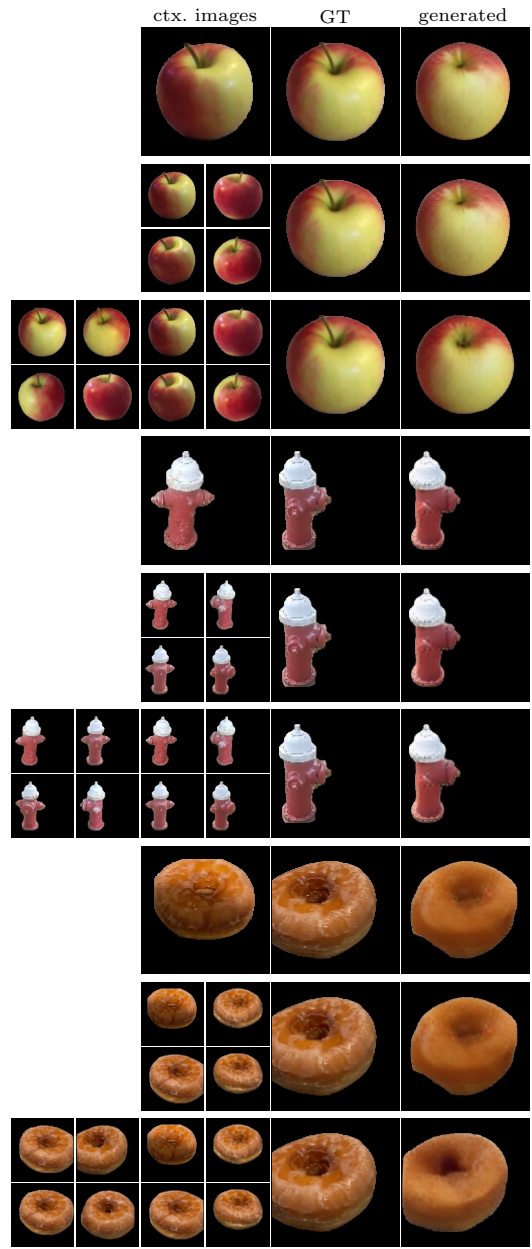

\begin{figure}[!t]
\centering
\begin{tikzpicture}[
 image/.style = {text width=0.132\linewidth, 
                 inner sep=0pt, outer sep=0pt},
label/.style = { minimum height=0.4cm },
node distance = 1pt and 1pt
                        ] 
\scriptsize
\path coordinate(last);
\node [image,below=of last,alias=last] (img00)
{\includegraphics[width=\linewidth]{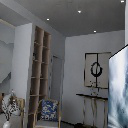}};
\node [image,right=of img00] (img01)
{\includegraphics[width=\linewidth]{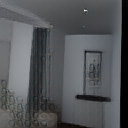}};
\node [image,right=1cm of img01] (img02) 
{\includegraphics[width=\linewidth]{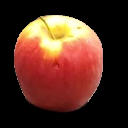}};
\node [image,right=of img02] (img03) 
{\includegraphics[width=\linewidth]{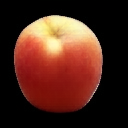}};

\node [image,below=of last,alias=last] (img10)
{\includegraphics[width=\linewidth]{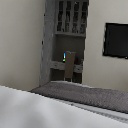}};
\node [image,right=of img10] (img11)
{\includegraphics[width=\linewidth]{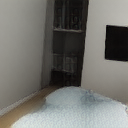}};
\node [image,right=1cm of img11] (img12) 
{\includegraphics[width=\linewidth]{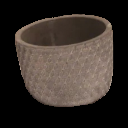}};
\node [image,right=of img12] (img13) 
{\includegraphics[width=\linewidth]{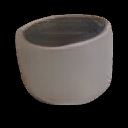}};

\node [image,below=of last,alias=last] (img10)
{\includegraphics[width=\linewidth]{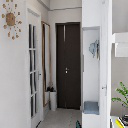}};
\node [image,right=of img10] (img11)
{\includegraphics[width=\linewidth]{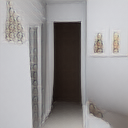}};
\node [image,right=1cm of img11] (img12) 
{\includegraphics[width=\linewidth]{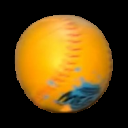}};
\node [image,right=of img12] (img13) 
{\includegraphics[width=\linewidth]{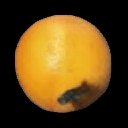}};

\node [image,below=of last,alias=last] (img10)
{\includegraphics[width=\linewidth]{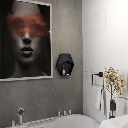}};
\node [image,right=of img10] (img11)
{\includegraphics[width=\linewidth]{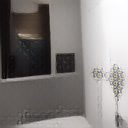}};
\node [image,right=1cm of img11] (img12) 
{\includegraphics[width=\linewidth]{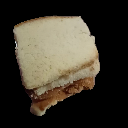}};
\node [image,right=of img12] (img13) 
{\includegraphics[width=\linewidth]{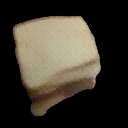}};

\node [image,below=of last,alias=last] (img10)
{\includegraphics[width=\linewidth]{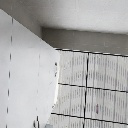}};
\node [image,right=of img10] (img11)
{\includegraphics[width=\linewidth]{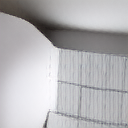}};
\node [image,right=1cm of img11] (img12) 
{\includegraphics[width=\linewidth]{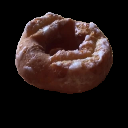}};
\node [image,right=of img12] (img13) 
{\includegraphics[width=\linewidth]{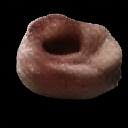}};

\node [image,below=of last,alias=last] (img10)
{\includegraphics[width=\linewidth]{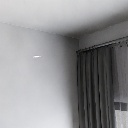}};
\node [image,right=of img10] (img11)
{\includegraphics[width=\linewidth]{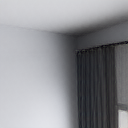}};
\node [image,right=1cm of img11] (img12) 
{\includegraphics[width=\linewidth]{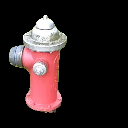}};
\node [image,right=of img12] (img13) 
{\includegraphics[width=\linewidth]{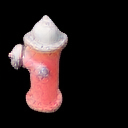}};

\node [image,below=of last,alias=last] (img10)
{\includegraphics[width=\linewidth]{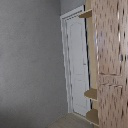}};
\node [image,right=of img10] (img11)
{\includegraphics[width=\linewidth]{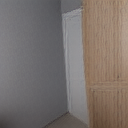}};
\node [image,right=1cm of img11] (img12) 
{\includegraphics[width=\linewidth]{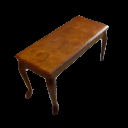}};
\node [image,right=of img12] (img13) 
{\includegraphics[width=\linewidth]{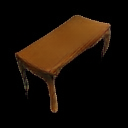}};

\node [image,below=of last,alias=last] (img10)
{\includegraphics[width=\linewidth]{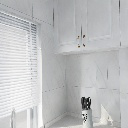}};
\node [image,right=of img10] (img11)
{\includegraphics[width=\linewidth]{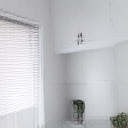}};
\node [image,right=1cm of img11] (img12) 
{\includegraphics[width=\linewidth]{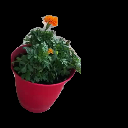}};
\node [image,right=of img12] (img13) 
{\includegraphics[width=\linewidth]{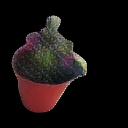}};

\node [image,below=of last,alias=last] (img10)
{\includegraphics[width=\linewidth]{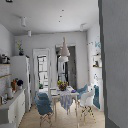}};
\node [image,right=of img10] (img11)
{\includegraphics[width=\linewidth]{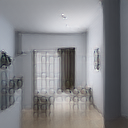}};
\node [image,right=1cm of img11] (img12) 
{\includegraphics[width=\linewidth]{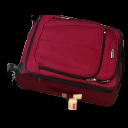}};
\node [image,right=of img12] (img13) 
{\includegraphics[width=\linewidth]{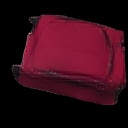}};

\node [image,below=of last,alias=last] (img10)
{\includegraphics[width=\linewidth]{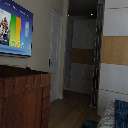}};
\node [image,right=of img10] (img11)
{\includegraphics[width=\linewidth]{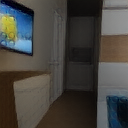}};
\node [image,right=1cm of img11] (img12) 
{\includegraphics[width=\linewidth]{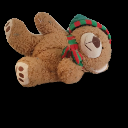}};
\node [image,right=of img12] (img13) 
{\includegraphics[width=\linewidth]{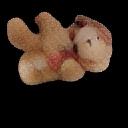}};

\node [label,above=-1pt of img00] {GT};
\node [label,above=-1pt of img01] {generated};
\node [label,above=-1pt of img02] {GT};
\node [label,above=-1pt of img03] {generated};
\end{tikzpicture}

\caption{Images generated on the InteriorNet dataset (\textbf{left}) with context size 19 and the CO3D dataset (\textbf{right}) with context size 9. For the CO3D evaluation, we used the model trained on all categories}
\label{fig:apx_interiornet_co3d}
\end{figure}

We add qualitative results to the ones presented in the paper (see Fig.~\ref{fig:teaser}, \ref{fig:interiornet}, and~\ref{fig:co3d} in the main paper). We show the context views together with the rendered images on the InteriorNet \cite{li2018interiornet}, the Common Objects in 3D (CO3D) \cite{reizenstein2021common}, and the 7-Scenes \cite{glocker2013real} datasets. The generated images are displayed in \cref{fig:apx_interiornet}, \cref{fig:apx_co3d}, and \cref{fig:apx_imggen-7scenes}, respectively. We also show images generated with full context sizes in \cref{fig:apx_interiornet_co3d}.
It is important to note that all the visualizations, including the video, were rendered on previously unseen scenes (objects).

The images rendered on the largest and most complex dataset -- InteriorNet, although slightly blurry, resemble the ground truth (GT) images well.
For the 7-Scenes dataset, the trained model overfitted the data, and the quality of the generated images was not as good as on other datasets.
Notice how the image rendered on CO3D is smoother than the ground truth image. In the case of the flower pot (\cref{fig:apx_co3d}), we can see that the model could not represent the particular shape and used a simpler shape instead. This is an intriguing property of the model which in the case of incomplete information uses its large prior to achieve more realistic renderings at the cost of being less similar to the real object.

\section{Attached video}\label{sec:attached_video}
We attach a video file\footnote{\url{https://jkulhanek.com/viewformer/video.html}} showing the generated images on various datasets. The video contains the results generated on the ShapeNet, CO3D, InteriorNet, and 7-Scenes datasets.
On the ShapeNet dataset, we compare our model with PixelNeRF \cite{yu2021pixelnerf}. We render video sequences of rotating objects using the same three context views.
For the CO3D dataset, we show video sequences of rotating objects using 9 context views. We also show how the model changes its prediction given more context views.
Unfortunately, we cannot compare with PixelNeRF \cite{yu2021pixelnerf} because the method was not able to converge properly on the dataset (see \cref{sec:experiments} in the main paper). Also, we cannot compare with NerFormer \cite{reizenstein2021common} because the source code is not publicly available.
Finally, we show the results on the InteriorNet dataset as well as on all scenes from the 7-Scenes dataset.

One might expect that with the discrete codebook codes the learned representation would be quantized and an arbitrary pose could not be represented by the model. However, from the sequences generated on the ShapeNet dataset, we can see that 
this problem does not occur and the model is able to capture the motion, smoothly transitioning between the true poses.
Therefore, although the codes are discrete, they can represent a continuous range of objects' orientations and positions.
It is interesting to see that our approach is occasionally not color consistent from frame to frame, \eg, see the police car at time 0:07.
We believe that the cause of this problem may stem from the codebook.
It was trained using a perceptual loss, which might be less sensitive to colors \cite{engilberge2017color}.
On the InteriorNet dataset (time 3:02), look at the pictures on the wall. The model first generates a window in place of the pictures, and with more context views, it replaces the window with two pictures. This illustrates well how the model improves its prediction given more context views.

\section{7-Scenes evaluation}\label{sec:7scenes}
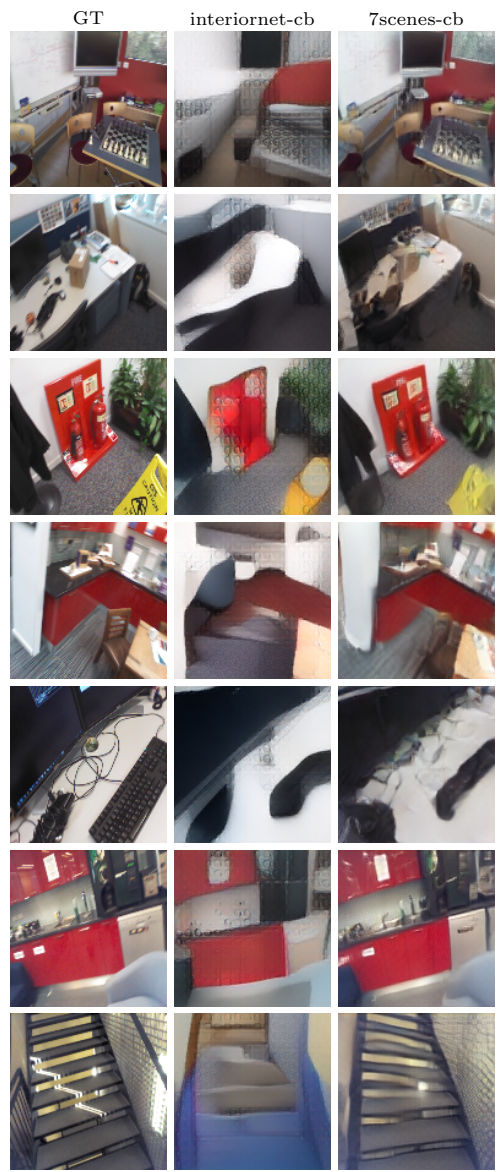
\begin{figure}[!t]
\centering
\begin{tikzpicture}[
 image/.style = {text width=0.17\linewidth, 
                 inner sep=0pt, outer sep=0pt},
label/.style = { minimum height=0.4cm },
node distance = 3pt and 3pt
                        ] 
\scriptsize
\path coordinate(last);
\node [image,below=of last,alias=last] (img00)
{\includegraphics[width=\linewidth,height=\linewidth]{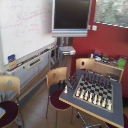}};
\node [image,right=of img00] (img01)
{\includegraphics[width=\linewidth]{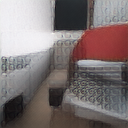}};
\node [image,right=of img01] (img02) 
{\includegraphics[width=\linewidth]{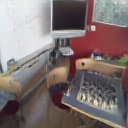}};

\node [image,below=of last,alias=last] (img10)
{\includegraphics[width=\linewidth,height=\linewidth]{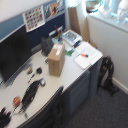}};
\node [image,right=of img10] (img11)
{\includegraphics[width=\linewidth]{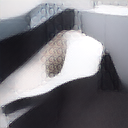}};
\node [image,right=of img11] (img12) 
{\includegraphics[width=\linewidth]{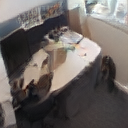}};

\node [image,below=of last,alias=last] (img10)
{\includegraphics[width=\linewidth,height=\linewidth]{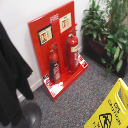}};
\node [image,right=of img10] (img11)
{\includegraphics[width=\linewidth]{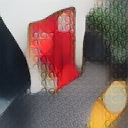}};
\node [image,right=of img11] (img12) 
{\includegraphics[width=\linewidth]{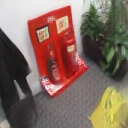}};

\node [image,below=of last,alias=last] (img10)
{\includegraphics[width=\linewidth,height=\linewidth]{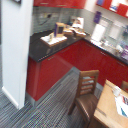}};
\node [image,right=of img10] (img11)
{\includegraphics[width=\linewidth]{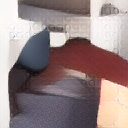}};
\node [image,right=of img11] (img12) 
{\includegraphics[width=\linewidth]{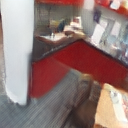}};

\node [image,below=of last,alias=last] (img10)
{\includegraphics[width=\linewidth,height=\linewidth]{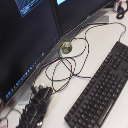}};
\node [image,right=of img10] (img11)
{\includegraphics[width=\linewidth]{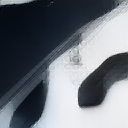}};
\node [image,right=of img11] (img12) 
{\includegraphics[width=\linewidth]{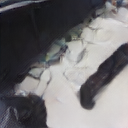}};

\node [image,below=of last,alias=last] (img10)
{\includegraphics[width=\linewidth,height=\linewidth]{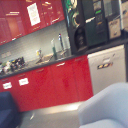}};
\node [image,right=of img10] (img11)
{\includegraphics[width=\linewidth]{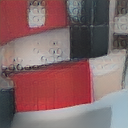}};
\node [image,right=of img11] (img12) 
{\includegraphics[width=\linewidth]{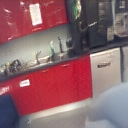}};

\node [image,below=of last,alias=last] (img10)
{\includegraphics[width=\linewidth,height=\linewidth]{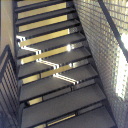}};
\node [image,right=of img10] (img11)
{\includegraphics[width=\linewidth]{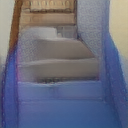}};
\node [image,right=of img11] (img12) 
{\includegraphics[width=\linewidth]{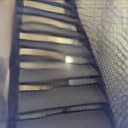}};
\node [label,above=-1pt of img00] {GT};
\node [label,above=-1pt of img01] {interiornet-cb};
\node [label,above=-1pt of img02] {7scenes-cb};
\end{tikzpicture}
\vspace{5mm}
\caption{Evaluation of the transformer model on the 7-Scenes dataset \cite{glocker2013real}. We display the ground-truth image (\textbf{GT}), the image generated using a codebook trained only on the InteriorNet dataset (\textbf{interiornet-cb}) and the image generated by a model with codebook fine-tuned on the 7-Scenes (\textbf{7scenes-cb}). For the visualization the context size was set to 19
\label{fig:apx_imggen-7scenes}}
\end{figure}

\begin{table*}[t]
\caption{
Camera pose estimation accuracy on the 7-Scenes dataset \cite{glocker2013real}, reported as the mean median position (in meters) and orientation (in degrees) errors over all scenes. We report results with an InteriorNet pre-trained codebook (`-in') and a codebook fine-tuned on 7-Scenes (`-7s'). We further compare a simple decoding scheme (random context views) with a variant that uses the top-10 most similar training images for each query view (`top10'), identified via image retrieval
\label{tab:7scenes}}
\setlength\tabcolsep{0.05cm}
\newcommand\hcline{\cmidrule{1-1} \cmidrule{2-9}}
\tiny\centering
\begin{tabular}{
@{}
 l@{\extracolsep{0.1cm}}
D{/}{/}{4.5}@{\extracolsep{0.1cm}}
D{/}{/}{4.4}@{\extracolsep{0.1cm}}
D{/}{/}{4.4}@{\extracolsep{0.1cm}}
D{/}{/}{4.4}@{\extracolsep{0.1cm}}
D{/}{/}{4.4}@{\extracolsep{0.1cm}}
D{/}{/}{4.4}@{\extracolsep{0.1cm}}
D{/}{/}{4.4}@{\extracolsep{0.1cm}}
D{/}{/}{4.5}@{\extracolsep{0.1cm}}
@{}
}
& \multicolumn{1}{c}{All} & \multicolumn{1}{c}{Chess} & \multicolumn{1}{c}{Fire}   & \multicolumn{1}{c}{Heads}  & \multicolumn{1}{c}{Office} & \multicolumn{1}{c}{Pumpkin} & \multicolumn{1}{c}{Kitchen} & \multicolumn{1}{c}{Stairs} \\
Method 
 & \multicolumn{1}{c}{Pos/Ori}
 & \multicolumn{1}{c}{Pos/Ori}
 & \multicolumn{1}{c}{Pos/Ori}
 & \multicolumn{1}{c}{Pos/Ori}
 & \multicolumn{1}{c}{Pos/Ori}
 & \multicolumn{1}{c}{Pos/Ori}
 & \multicolumn{1}{c}{Pos/Ori}
 & \multicolumn{1}{c}{Pos/Ori} \\
 \hcline
\Ours-in%
& 0.24/10.49   & 0.16/8.03    & 0.24/11.35   & 0.17/13.23   & 0.25/10.33   & 0.23/8.20    & 0.31/11.01   & 0.30/11.28  \\
\Ours-in-top10 %
& 0.19/7.82    & 0.13/6.36    & 0.22/10.27   & 0.17/10.85   & 0.17/6.42    & 0.19/6.26    & 0.21/6.62    & 0.21/7.97   \\
\Ours-7s%
& 0.23/8.46    & 0.15/6.31    & 0.23/10.03   & 0.19/12.68   & 0.23/7.69    & 0.19/5.59    & 0.27/7.75    & 0.31/9.18 \\
\Ours-7s-top10%
& 0.17/6.68    & 0.12/4.85    & 0.20/8.65    & 0.17/10.41   & 0.15/5.11    & 0.16/4.78    & 0.18/5.01    & 0.22/7.93 \\
\cmidrule{1-1} \cmidrule{2-9}
Oracle-top10          & 0.21/10.01   & 0.18/9.16   & 0.27/10.37   & 0.12/11.44  & 0.22/8.33   & 0.24/8.20   & 0.26/9.72   & 0.19/12.85  \\
\cmidrule{1-1} \cmidrule{2-9}
PoseNet \cite{kendall2015posenet} & 0.44/10.4 & 0.32/8.12 & 0.47/14.4 & 0.29/12.0 & 0.48/7.68 & 0.47/8.42 & 0.59/8.64 & 0.47/13.8 \\
MapNet \cite{brahmbhatt2018geometry} & 0.18/6.56 & 0.09/3.24 & 0.20/9.29 & 0.12/8.45 & 0.19/5.45 & 0.19/3.96 & 0.20/4.94 & 0.27/10.57 \\

LENS \cite{moreau2021lens} & 0.05/2.5 & 0.04/2.0 & 0.03/1.5 & 0.02/1.5 &  0.09/3.6 & 0.08/3.1 & 0.07/3.4 & 0.03/2.2 \\

MS-Transformer \cite{shavit2021learning} &  0.18/7.28 & 0.11/4.66 & 0.24/9.6  & 0.14/12.19 & 0.17/5.66 & 0.18/4.44 & 0.17/5.94 & 0.26/8.45 \\

RelocNet \cite{balntas2018relocnet} & 0.21/6.72 & 0.12/4.14 & 0.26/10.4 & 0.14/10.5 & 0.18/5.32 & 0.26/4.17 & 0.23/5.0 & 0.28/7.53 \\

CamNet \cite{ding2019camnet} & 0.04/1.69 & 0.04/1.73 & 0.03/1.74 &  0.05/1.98 &  0.04/1.62 &  0.04/1.64 &  0.04/1.63 &  0.04/1.51  \\

DenseVLAD \cite{sattler2019understanding,torii201524} & 0.26/13.1 & 0.21/12.5 & 0.33/13.8 & 0.15/14.9 & 0.28/11.2 & 0.31/11.3 & 0.30/12.3 & 0.25/15.8 \\
DenseVLAD+Int.\cite{sattler2019understanding} & 0.24/11.7 & 0.18/10.0 & 0.33/12.4 & 0.14/14.3 & 0.25/10.1 & 0.26/9.42 & 0.27/11.1 & 0.24/14.7 \\
\cmidrule{1-1} \cmidrule{2-9}

DSAC* \cite{brachmann2020ARXIV} & 0.03/1.36 & 0.02/1.10 & 0.02/1.24 & 0.01/1.82 & 0.03/1.15 & 0.04/1.34 & 0.04/1.68 & 0.03/1.16 \\
hloc \cite{sarlin2019cvpr} & 0.03/1.09 & 0.02/0.85  & 0.02/0.94  & 0.01/0.75  & 0.03/0.92  & 0.05/1.30  & 0.04/1.40 & 0.05/1.47 \\
Active Search \cite{sattler2016efficient} & 0.04/1.18
 & 0.03/0.87  & 0.02/1.01  & 0.01/0.82  & 0.04/1.15  & 0.07/1.69  & 0.05/1.72  & 0.04/1.01 \\

\bottomrule
\end{tabular}

\end{table*}

In order to evaluate the performance of our approach on the task of camera pose estimation, we present the results on a localization benchmark dataset -- 7-Scenes \cite{glocker2013real} (\cf \cref{sec:experiments} in the main paper). We trained two models -- one with a fine-tuned codebook and the other one with the InteriorNet-trained codebook. For all models, we used context size 19.
We have evaluated the method on all views from the test set of each of the 7 scenes and used the views from the training set as context images. Generated images can be seen in \cref{fig:apx_imggen-7scenes}.

For localization, we have experimented with different strategies for obtaining the context view required by our approach: 
by default, we simply randomly select 19 training images as context for each test image. 
We further evaluate a variant that uses the top-10 most similar images identified via image retrieval with DenseVLAD~\cite{torii201524} descriptors (indicated as ``-top10").  
The remaining 9 context images are randomly selected from the training images. We also experimented with using the top-19 retrieved images but found this approach to work worse. We attribute this to the fact that the images of the 7-Scenes datasets are taken in sequences and that there is little viewpoint variation between the top-19 retrieved images.

We evaluate variants where the codebook is trained only on InteriorNet (indicated as ``-in") and where the codebook is fine-tuned on the training images of 7-Scenes (``-7s``). 
As can be seen in \cref{tab:7scenes}, using a fine-tuned codebook improves performance. 
Similarly, using the top-10 retrieved images leads to more accurate camera poses. 
For evaluation, we follow the common practice and report the median position and orientation error per scene, as well as the mean median position and mean median orientation error over all the scenes.

To better understand the performance of our approach, we compare it against an oracle. 
Given the top-10 retrieved images via DenseVLAD, the oracle selects the retrieved image with the smallest position and the smallest orientation error. 
As shown in~\cref{tab:7scenes}, our approach outperforms the oracle on most scenes. This implies that the model is able to interpolate the context views such that it generates a pose that is closer to the query than any other in the context. 

\cref{tab:7scenes} also includes comparison with various baselines. 
Absolute pose regression techniques~\cite{brahmbhatt2018geometry,kendall2015posenet,moreau2021lens,shavit2021learning} train a CNN to directly regress the camera pose for a given input image. 
Our approach performs similarly well or better than these baselines, with the exception of LENS~\cite{moreau2021lens}, which uses additional training data in the form of images rendered from novel viewpoints. 
Our approach also typically outperforms the two image retrieval-based baselines (DenseVLAD and DenseVLAD + Int.) They were proposed in~\cite{sattler2019understanding} as a form of sanity check for absolute pose regression approaches. 

Similar to our approach, relative pose regression approaches~\cite{balntas2018relocnet,ding2019camnet} estimate the pose of the test image \wrt a set of context views. 
These context views are obtained by finding the most similar training images using image retrieval. 
Our approach performs similarly well (and often better) as RelocNet~\cite{balntas2018relocnet}, which also uses a single forward pass to regress relative poses (between pairs of images). 
CamNet~\cite{ding2019camnet} uses a more complicated pipeline consisting of coarse and fine relative pose regression stages, which results in higher accuracy. 

Structure-based approaches use 2D-3D matches between pixels in a test image and 3D scene points~\cite{Brachmann2021TPAMI,sarlin2019cvpr,sattler2016efficient}. 
These approaches currently represent the state-of-the-art in terms of pose accuracy and are more accurate than pose regression-based techniques. 
In contrast to the other baselines, they store the 3D structure of the scene.
Overall, the results show that our approach achieves a similar level of pose accuracy as comparable methods.

\section{Ablation study}\label{sec:ablation_study}
We compare our model with alternative architectures to validate the design choices we made. We also demonstrate how the quality of predictions improves with larger context sizes. The InteriorNet dataset \cite{li2018interiornet} was used for all evaluations because of its large size. The context size was 19.

\noindent\textbf{Different model variants.}
We compare variants of our approach trained for only one of the two tasks -- image generation and localization -- on the InteriorNet dataset \cite{li2018interiornet}. We also evaluate the importance of the proposed branching attention by training alternative language models (LMs) that do not use it. As discussed in \cref{sec:introduction} in the main paper, one way to train the transformer without the branching attention is to have a purely autoregressive (causal) LM \cite{radford2019language,vaswani2017attention}. These models were successfully applied to similar tasks \cite{esser2021taming,parmar2018image,ramesh2021zero}. We also train another alternative -- masked LMs -- that benefits from the same inference speed as our method \cite{devlin2019bert}.
In particular, the following models are compared:
\begin{itemize}
\item \textbf{ViewFormer} -- our approach with both localization and image generation enabled.
\item \textbf{ViewFormer no-loc} -- our approach without localization.
\item \textbf{ViewFormer no-imagen} -- our approach without image generation.
\item \textbf{Causal LM} -- the same transformer model with autoregressive decoding. Instead of decoding all tokens at once, we model the probability distribution over the next image token given all previous tokens \cite{radford2019language,vaswani2017attention}.
\item \textbf{Causal LM + masked loc.} -- causal LM with added localization. For the localization, we mask the poses of three random views from the training batch and attach a regression head to the last token of each image.
\item \textbf{Masked LM} -- the same transformer model with masked decoding (without the branching attention). We randomly mask three views from the training sequence and train the model to recover it. Note that the model is optimized for a single context size (previous variants optimized for all context sizes).
\item \textbf{Masked LM + masked loc.} -- masked LM with added localization. For the localization, we mask the poses of three random views from the training batch and attach a regression head to all image tokens. The resulting poses are averaged in the same way as in ViewFormer.
\end{itemize}

\begin{figure}[!t]
    \centering
\newcommand\aqDataRow[1]{
    \node [image,below=of last,alias=last] (img0)
    {\includegraphics[width=\linewidth]{resources/ablations/gt/#1.png}};
    \node [image,right=4pt of img0] (img1)
    {\includegraphics[width=\linewidth]{resources/ablations/viewformer/#1.png}};
    \node [image,right=of img1] (img2) 
    {\includegraphics[width=\linewidth]{resources/ablations/viewformer-noloc/#1.png}};
    \node [image,right=of img2] (img3) 
    {\includegraphics[width=\linewidth]{resources/ablations/gpt2-noloc/#1.png}};
    \node [image,right=of img3] (img4) 
    {\includegraphics[width=\linewidth]{resources/ablations/gpt2-loc/#1.png}};
    \node [image,right=of img4] (img5) 
    {\includegraphics[width=\linewidth]{resources/ablations/mgpt-noloc/#1.png}};
    \node [image,right=of img5] (img6) 
    {\includegraphics[width=\linewidth]{resources/ablations/mgpt-loc/#1.png}};
}
\begin{tikzpicture}[
 image/.style = {text width=0.132\linewidth, 
                 inner sep=0pt, outer sep=0pt},
label/.style = { minimum height=0.4cm },
node distance = 1pt and 1pt
                        ] 
\tiny
\path coordinate(last);
\aqDataRow{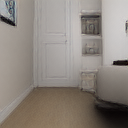};

\node [label,above=1pt of img0,align=center] {GT\\};
\node [label,above=1pt of img1,align=center] {ViewFormer\\};
\node [label,above=1pt of img2,align=center] {ViewFormer\\ no-loc};
\node [label,above=1pt of img3,align=center] {Causal LM\\};
\node [label,above=1pt of img4,align=center] {Causal LM\\ masked loc.};
\node [label,above=1pt of img5,align=center] {Masked LM\\};
\node [label,above=1pt of img6,align=center] {Masked LM\\ masked loc.};

\aqDataRow{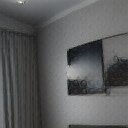};
\aqDataRow{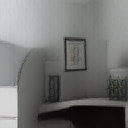};
\aqDataRow{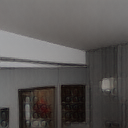};
\aqDataRow{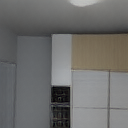};

\end{tikzpicture}

    \caption{Examples generated by alternative architectures described in \cref{sec:ablation_study}. The examples were generated on the test set of the InteriorNet dataset using context size 19.
        \label{fig:apx_ablation}
    }
\end{figure}
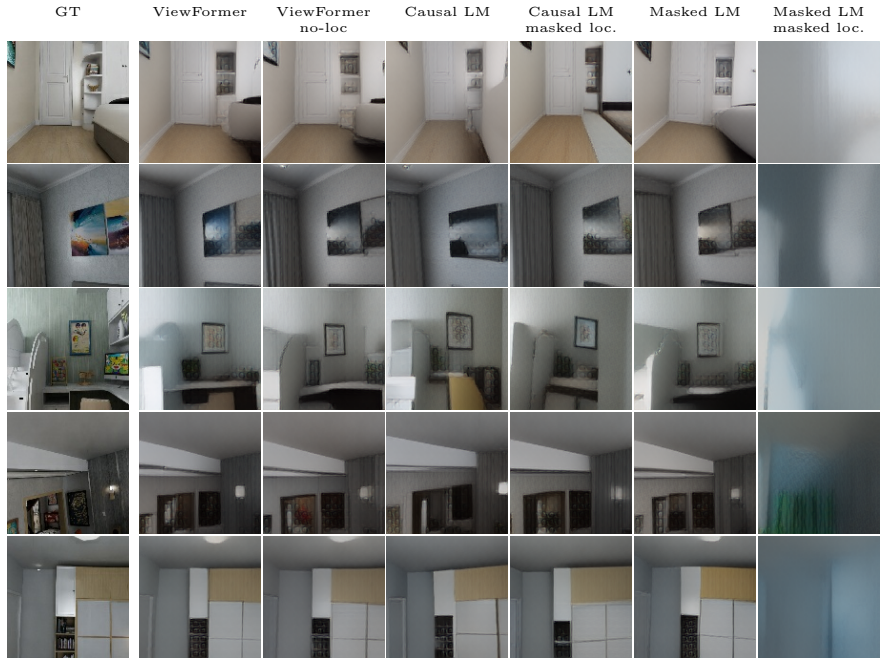

The results (averaged over all test scenes) are shown in \cref{tab:interiornet}. We also include a qualitative comparison in \cref{fig:apx_ablation}.
As can be seen, training without the localization task improves image quality, whereas there is little difference in terms of pose accuracy between training with or without the generation. 

Our method outperforms both causal LM and masked LM in image generation performance and localization accuracy. Note that our decoding is much faster compared to causal LM because we decode all tokens at once (see \Cref{sec:introduction} in the main paper). For a causal LM, generating a single view takes 10\,s even when using cache. Compare this to 93\,ms for the ViewFormer. Compared to masked LM, our model has the same inference speed, but the added benefit of being optimized for all context sizes. Masked LM can be optimized for one context size only.

\begin{table}[t!]
\caption{
Ablation study evaluated on the InteriorNet dataset \cite{li2018interiornet}. See \cref{sec:ablation_study} for a  description of the compared variants.
We show the PSNR, the pixel-wise MAE, and the LPIPS distance \cite{zhang2018unreasonable}. For localization, we show the median position error in meters and the median orientation error in degrees computed over all scenes.
\label{tab:interiornet}}
\setlength\tabcolsep{0.05cm}
\scriptsize\centering
\begin{tabular}{
@{}
l@{\extracolsep{0.1cm}}
ccc@{\extracolsep{0.1cm}}
c@{\extracolsep{0.1cm}}
@{}}
& \multicolumn{3}{c}{\textbf{Image generation}} & \multicolumn{1}{c}{\textbf{Localization}} \\
\cmidrule{2-4} \cmidrule{5-5}
Method & PSNR$\uparrow$ & MAE$\downarrow$ & LPIPS$\downarrow$ & Pos/Ori$\downarrow$ \\
\cmidrule{1-1} \cmidrule{2-4} \cmidrule{5-5}
\textbf{\ours} & 18.53  & 23.35  & 0.33 & \textbf{0.19}/\textbf{4.22} \\
\ours no-loc & \textbf{19.10}  & \textbf{21.56}  & \textbf{0.32} & - \\
\ours no-imagen & - & - & - & \textbf{0.19}/4.34 \\
\cmidrule{1-1} \cmidrule{2-5}
Causal LM & 16.75 & 29.88 & 0.39 & - \\
Causal LM + masked loc. & 16.67 & 30.22 & 0.39 & 0.22/6.24 \\
Masked LM & 18.76 & 22.91 & \textbf{0.32} & -\\
Masked LM + masked loc. & 14.51 & 42.89 & 0.51 & 0.32/29.65 \\
\bottomrule
\end{tabular}
\end{table}

\begin{figure}[t]
\subfloat{
\includegraphics[width=0.47\linewidth]{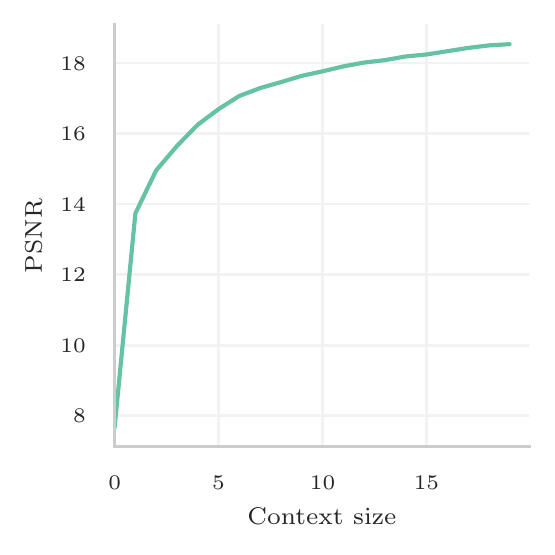}
}
\hfill
\subfloat{
\includegraphics[clip,width=0.47\linewidth]{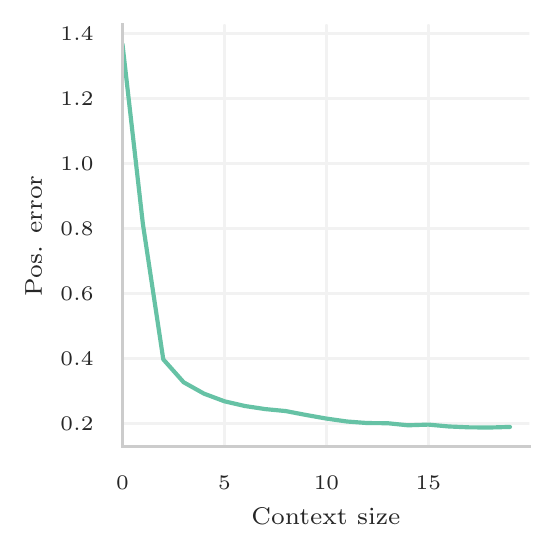}
}
\caption{This plot shows the effect of increasing the context size on the PSNR (\textbf{left}) and the position error (\textbf{right}) evaluated on the InteriorNet dataset \cite{li2018interiornet} 
\label{fig:apx_interiornet_multictx}}
\end{figure}

\noindent\textbf{Increasing the context size.}
We show the effect of increasing the context size on localization and image generation performance. The image generation performance (measured with PSNR) and the localization accuracy (median Euclidean distance between the predicted camera position and the ground truth) are shown in \cref{fig:apx_interiornet_multictx}. The results were computed on all scenes from the test set.

We can see that the performance of both novel view synthesis and camera pose estimation increases with more context views. The change is most prominent in the first five views, but after that it keeps increasing as well.

\section{ShapeNet evaluation}\label{sec:shapenet}
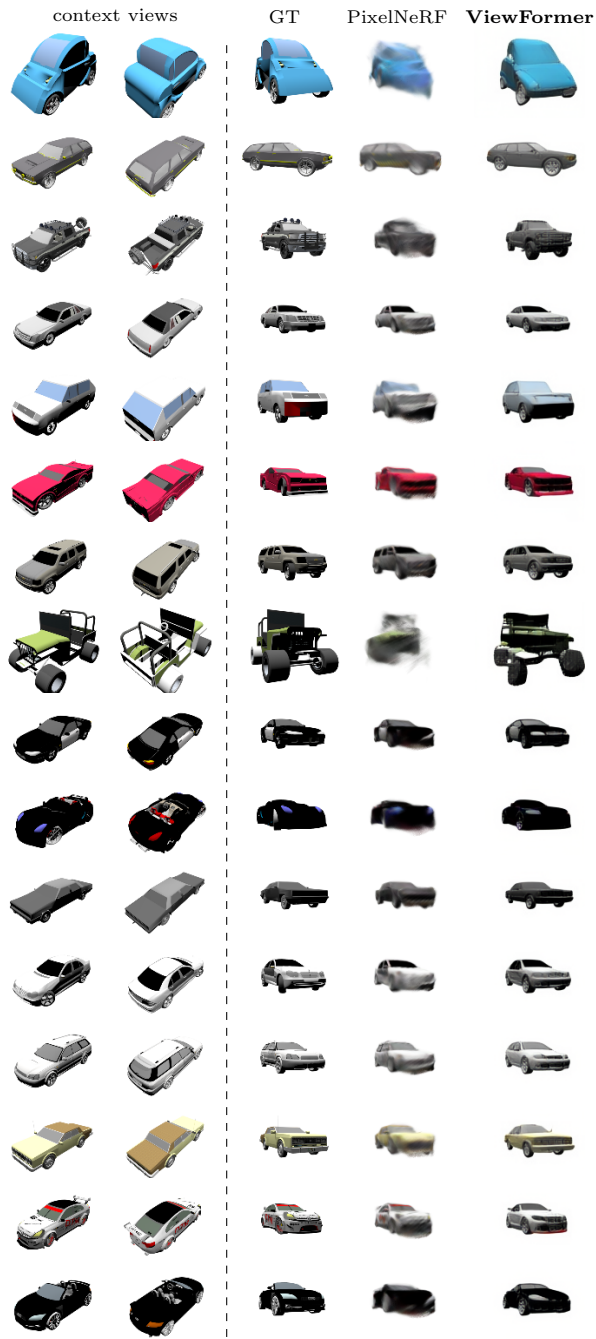
\begin{figure}[!p]
    \centering
    \newcommand\figrowwideb[2]{
    \node [image,below=of last,alias=last] (img0)
    {\includegraphics[width=\linewidth,trim={0 1mm 0 7mm},clip]{#1/gt/00-064.png}};
    \node [image,right=of img0] (img1)
    {\includegraphics[width=\linewidth,trim={0 1mm 0 7mm},clip]{#1/gt/01-104.png}};
    \node [image,right=of img1] (img2)
    {\includegraphics[width=\linewidth,trim={0 1mm 0 7mm},clip]{#1/gt/#2.png}};
    \node [image,right=of img2] (img3) 
    {\includegraphics[width=\linewidth,trim={0 1mm 0 7mm},clip]{#1/pixelnerf/000#2.png}};
    \node [image,right=0.3cm of img3] (img4) 
    {\includegraphics[width=\linewidth,trim={0 1mm 0 7mm},clip]{#1/viewformer/#2.png}};
}
\newcommand\figrowwide[2]{
    \node [image,below=of last,alias=last] (img0)
    {\includegraphics[width=\linewidth,trim={0 5mm 0 7mm},clip]{#1/gt/00-064.png}};
    \node [image,right=of img0] (img1)
    {\includegraphics[width=\linewidth,trim={0 5mm 0 7mm},clip]{#1/gt/01-104.png}};
    \node [image,right=of img1] (img2)
    {\includegraphics[width=\linewidth,trim={0 5mm 0 7mm},clip]{#1/gt/#2.png}};
    \node [image,right=of img2] (img3) 
    {\includegraphics[width=\linewidth,trim={0 5mm 0 7mm},clip]{#1/pixelnerf/000#2.png}};
    \node [image,right=0.3cm of img3] (img4) 
    {\includegraphics[width=\linewidth,trim={0 5mm 0 7mm},clip]{#1/viewformer/#2.png}};
}
\begin{tikzpicture}[
     image/.style = {text width=0.12\textwidth, 
                     inner sep=0pt, outer sep=0pt},
    label/.style = {  },
    node distance = 0pt and 1pt
                            ]

    \scriptsize
    \path coordinate(last);
    
    \figrowwideb{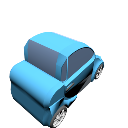}{000}
    \node [label,above=-0.5pt of $(img0.north)!0.5!(img1.north)$] {context views};
    \node [label,above=-1pt of img2] {GT};
    \node [label,above=-1pt of img3] {PixelNeRF};
    \node [label,above=-1pt of img4] {\textbf{ViewFormer}};
    \figrowwide{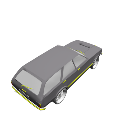}{008}

    \figrowwide{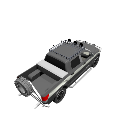}{000}
    \figrowwide{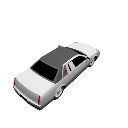}{000}
    \figrowwide{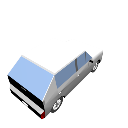}{000}
    \figrowwide{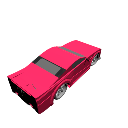}{000}
    \figrowwide{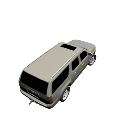}{000}
    \figrowwideb{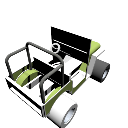}{000}
    \figrowwide{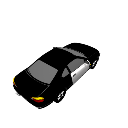}{000}
    \figrowwide{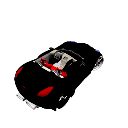}{000}
    \figrowwide{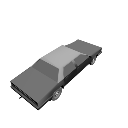}{000}
    \figrowwide{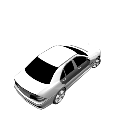}{000}
    \figrowwide{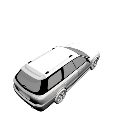}{000}
    \figrowwide{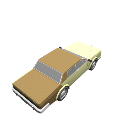}{000}
    \figrowwide{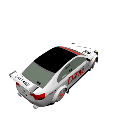}{000}
    \figrowwide{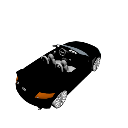}{000}

    \draw[dashed] ([yshift=16.15cm]img1.north east) -- ([yshift=0cm]img1.south east); 
    \end{tikzpicture}

    \caption{
    Additional \textbf{ShapeNet} cars qualitative comparison with PixelNeRF \cite{yu2021pixelnerf} using two context views
\label{fig:apx_shapenet_cars}
}
\end{figure}

\begin{figure}[!p]
    \centering
    \newcommand\figrowwideb[2]{
    \node [image,below=of last,alias=last] (img0)
    {\includegraphics[width=\linewidth,trim={0 1mm 0 7mm},clip]{#1/gt/00-064.png}};
    \node [image,right=of img0] (img1)
    {\includegraphics[width=\linewidth,trim={0 1mm 0 7mm},clip]{#1/gt/01-104.png}};
    \node [image,right=of img1] (img2)
    {\includegraphics[width=\linewidth,trim={0 1mm 0 7mm},clip]{#1/gt/#2.png}};
    \node [image,right=of img2] (img3) 
    {\includegraphics[width=\linewidth,trim={0 1mm 0 7mm},clip]{#1/pixelnerf/000#2.png}};
    \node [image,right=0.3cm of img3] (img4) 
    {\includegraphics[width=\linewidth,trim={0 1mm 0 7mm},clip]{#1/viewformer/#2.png}};
}
\newcommand\figrowwide[2]{
    \node [image,below=of last,alias=last] (img0)
    {\includegraphics[width=\linewidth,trim={0 5mm 0 7mm},clip]{#1/gt/00-064.png}};
    \node [image,right=of img0] (img1)
    {\includegraphics[width=\linewidth,trim={0 5mm 0 7mm},clip]{#1/gt/01-104.png}};
    \node [image,right=of img1] (img2)
    {\includegraphics[width=\linewidth,trim={0 5mm 0 7mm},clip]{#1/gt/#2.png}};
    \node [image,right=of img2] (img3) 
    {\includegraphics[width=\linewidth,trim={0 5mm 0 7mm},clip]{#1/pixelnerf/000#2.png}};
    \node [image,right=0.3cm of img3] (img4) 
    {\includegraphics[width=\linewidth,trim={0 5mm 0 7mm},clip]{#1/viewformer/#2.png}};
}
\begin{tikzpicture}[
     image/.style = {text width=0.12\textwidth, 
                     inner sep=0pt, outer sep=0pt},
    label/.style = {  },
    node distance = 0pt and 1pt
                            ]

    \scriptsize
    \path coordinate(last);
    
    \figrowwide{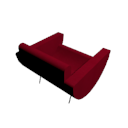}{000}
    \node [label,above=-0.5pt of $(img0.north)!0.5!(img1.north)$] {context views};
    \node [label,above=-1pt of img2] {GT};
    \node [label,above=-1pt of img3] {PixelNeRF};
    \node [label,above=-1pt of img4] {\textbf{ViewFormer}};
    \figrowwide{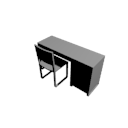}{000}

    \figrowwide{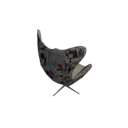}{000}
    \figrowwide{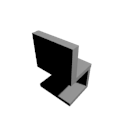}{000}
    \figrowwide{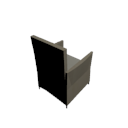}{000}
    \figrowwide{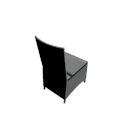}{000}
    \figrowwide{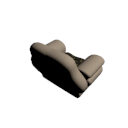}{000}
    \figrowwide{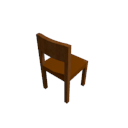}{000}
    \figrowwide{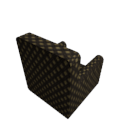}{000}
    \figrowwide{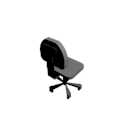}{000}
    \figrowwide{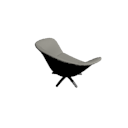}{000}
    \figrowwide{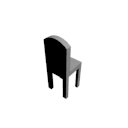}{000}
    \figrowwide{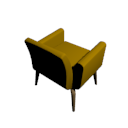}{000}
    \figrowwide{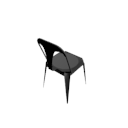}{000}
    \figrowwide{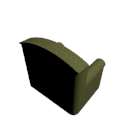}{000}
    \figrowwide{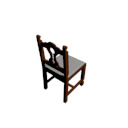}{000}
    
    \draw[dashed] ([yshift=16.0cm]img1.north east) -- ([yshift=0cm]img1.south east); 
    \end{tikzpicture}

    \caption{
    Additional \textbf{ShapeNet} chairs qualitative comparison with PixelNeRF \cite{yu2021pixelnerf} using two context views
\label{fig:apx_shapenet_chairs}
}
\end{figure}

In this section, we give more details on the ShapeNet results from the main paper (\cref{fig:shapenet}). We include quantitative and additional qualitative results.
We trained our model on ShapeNet dataset rendered by SRN \cite{sitzmann2019scene}. The context size used for training was three. We compare ViewFormer with SRN \cite{sitzmann2019scene} and PixelNeRF \cite{yu2021pixelnerf}. We show the PSNR and SSIM \cite{wang2009mean} averaged across color channels for both car and chair categories with one or two context views. The results are presented in \cref{tab:shapenet}. We also extend \cref{fig:shapenet} from the paper with additional qualitative results on cars and chairs in Fig.~\ref{fig:apx_shapenet_cars} and~\ref{fig:apx_shapenet_chairs}.

From the results, we can see that our method performs worse than both SRN \cite{sitzmann2019scene} and PixelNeRF \cite{yu2021pixelnerf} in terms of the quantitative results. This is expected because our method was designed for more views (more than 10) and was evaluated using one or two views. 
However, compared to PixelNeRF our method is able to recover more detail, whereas PixelNeRF produces blurry output, especially on the car category. Based on the qualitative results, we argue that although our approach has worse quantitative numbers, our results look more realistic. A possible cause for this observation could be that blurring the edges of an object can hide the unprecise geometry rendered by the model and increase PSNR. However, it loses fine detail in the images.

\begin{table}[!t]
    \caption{ShapeNet results comparing ViewFormer with SRN \cite{sitzmann2019scene} and PixelNeRF \cite{yu2021pixelnerf}. We show the results for both car and chair category with one or two context views
    \label{tab:shapenet}}
    \centering
    \newcommand{\method}[1]{{\oneptsmaller{\sffamily #1}}}
\newcommand{\methodrbtl}[1]{{\small{\sffamily #1}}}

\newcommand{\bb}[1]{{\textbf{\underline{#1}}}}
\renewcommand{\b}[1]{{\underline{#1}}}

\setlength\tabcolsep{0.05cm}
\scriptsize\centering
\newcommand{\isad}{\textsuperscript{\textdagger}}
\begin{tabular}{
@{}
lr@{\extracolsep{0.1cm}}
cc@{\extracolsep{0.1cm}}
cc@{\extracolsep{0.1cm}}
cc@{\extracolsep{0.1cm}}
cc
@{}
}%
               & & \multicolumn{2}{c}{cars 1 view} 
               & \multicolumn{2}{c}{cars 2 views} 
               & \multicolumn{2}{c}{chairs 1 view}
               & \multicolumn{2}{c}{chairs 2 views}\\
               \cmidrule{3-6} \cmidrule{7-10}

                Method & 3D & PSNR$\uparrow$ & SSIM$\uparrow$ & PSNR$\uparrow$ & SSIM$\uparrow$ & 
                PSNR$\uparrow$ & SSIM$\uparrow$ & 
                PSNR$\uparrow$ & SSIM$\uparrow$ 
                \\ \cmidrule{1-2} \cmidrule{3-4} \cmidrule{5-6} \cmidrule{7-8} \cmidrule{9-10}
                
 \textbf{\method{\Ours}} & \xmark & 19.03 & 0.83 & 20.09 & 0.85 & 14.74 & 0.79 & 17.20 & 0.84 \\
 \cmidrule{1-2} \cmidrule{3-10}
 
\method{SRN \cite{sitzmann2019scene}} & \cmark  & 22.25 & 0.89 & 24.84 & 0.92 & 22.89 & 0.89 & 24.48 & 0.92 \\
\method{PixelNeRF \cite{yu2021pixelnerf}} & \cmark & 23.72 & 0.91 & 26.20 & 0.94 & 23.17 & 0.90 & 25.66 & 0.94 \\
\bottomrule
\end{tabular}%

\end{table}

\section{Shepard-Metzler-Parts-7 evaluation}\label{sec:sm7_evaluation}
\begin{figure}[t]
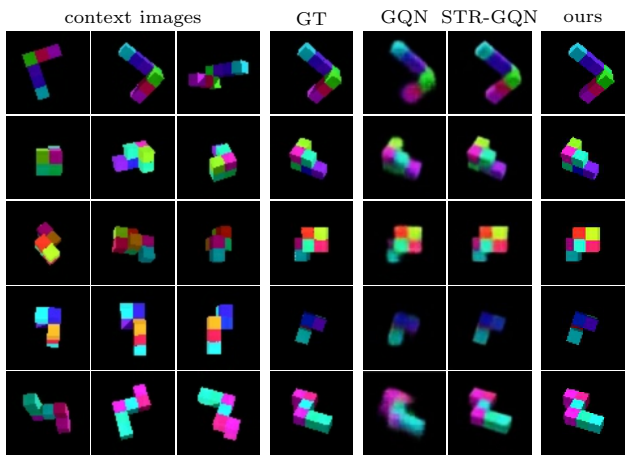

\centering
\begin{tikzpicture}[
 image/.style = {text width=0.09\linewidth, 
                 inner sep=0pt, outer sep=0pt},
label/.style = { minimum height=0.4cm },
node distance = 1pt and 1pt
                        ] 
\scriptsize
\path coordinate(last);
\foreach \i in {0,1,...,4} {
    \node [image,below=of last,alias=last] (img\i0)
    {\includegraphics[width=\linewidth]{resources/gqn/c\i_0.png}};
\node [image,right=of img\i0] (img\i1)
    {\includegraphics[width=\linewidth]{resources/gqn/c\i_1.png}};
\node [image,right=of img\i1] (img\i2) 
    {\includegraphics[width=\linewidth]{resources/gqn/c\i_2.png}};
    \node [image,right=4pt of img\i2] (img\i3) 
    {\includegraphics[width=\linewidth]{resources/gqn/gt\i.png}};
    \node [image,right=4pt of img\i3] (img\i4) 
    {\includegraphics[width=\linewidth]{resources/gqn/gqn\i.png}};
    \node [image,right=of img\i4] (img\i5) 
    {\includegraphics[width=\linewidth]{resources/gqn/str-gqn\i.png}};
    \node [image,right=4pt of img\i5] (img\i6)
    {\includegraphics[width=\linewidth]{resources/gqn/gen\i.png}};
}
\node [label,above=-1pt of img01] {context images};
\node [label,above=-1pt of img03] {GT};
\node [label,above=-1pt of img04] {GQN};
\node [label,above=-1pt of img05] {STR-GQN};
\node [label,above=-1pt of img06] {ours};
\end{tikzpicture}
\caption{Qualitative results on the SM7 dataset \cite{eslami2018neural}.
We compare against GQN \cite{eslami2018neural} and STR-GQN \cite{chen2021str}
\label{fig:apx_migt-sm7}}
\end{figure}

We evaluated our model on the Shepard-Metzler-Parts-7 dataset  \cite{eslami2018neural,shepard1971mental} to compare our approach to other methods that only operate in 2D~\cite{chen2021str,eslami2018neural,tobin2019geometry}. 
For the evaluation, we used the context size three.
The additional qualitative results, presented in \cref{fig:apx_migt-sm7}, extend \cref{fig:migt-sm7} from the main paper. Unfortunately, in the qualitative analysis, we cannot compare with E-GQN \cite{tobin2019geometry} because the authors did not make the generated images or models public.

\begin{table}[t]
\caption{Comparison with GQN-based methods \cite{chen2021direct,eslami2018neural,tobin2019geometry} on the SM7 dataset.
We show the \textbf{MAE}, \textbf{RMSE}, and the position and orientation errors (\textbf{Pos},  \textbf{Ori})\label{tab:migt-sm7}}
\setlength\tabcolsep{0.05cm}
\scriptsize\centering
\newcommand{\method}[1]{#1}
\begin{tabular}{
@{}
l@{\extracolsep{0.1cm}}
cc@{\extracolsep{0.1cm}}
c@{\extracolsep{0.1cm}}
@{}}
& \multicolumn{2}{c}{\bf{Image generation}} & \multicolumn{1}{c}{\bf{Localization}}\\
\cmidrule{2-3} \cmidrule{4-4}
Method & MAE$\downarrow$ & RMSE$\downarrow$ & Pos/Ori$\downarrow$\\
\cmidrule{1-1} \cmidrule{2-3} \cmidrule{4-4}
\textbf{\method{\ours}} & \bf{1.61} & 7.02 & 0.21/3.48 \\
\method{GQN \cite{eslami2018neural}} & 3.13 &9.97 & - \\
\method{E-GQN \cite{tobin2019geometry}} & 2.14 & \bf{5.63} & - \\
\method{STR-GQN \cite{chen2021direct}} & 3.11 & 10.56  & - \\
\hline
\end{tabular}
\end{table}
 \cref{tab:migt-sm7} presents quantitative results (averaged over 1000 scenes). As our method uses images of sizes $128 \times 128$ pixels, we rescaled the images before training the codebook. For evaluation, we used the original image size $64 \times 64$ pixels of the dataset.
We report the pixel-wise mean absolute error (MAE) and root mean square error (RMSE). 
For reference, we also show the localization accuracy.
The position error (Pos) is the median euclidean distance between the predicted positions and the ground-truth camera positions, and the orientation error (Ori) is the median of the angular distances in degrees.

As can be seen, our method clearly outperforms the baselines in terms of the MAE. 
E-GQN performs best in terms of the RMSE as it is trained to optimize this metric, whereas our method uses MAE and perceptual loss. 

\section{Codebook evaluation}\label{sec:codebook_evaluation}
\begin{table}[t]
\caption{Codebook evaluation on the SM7 \cite{eslami2018neural,shepard1971mental}, InteriorNet \cite{li2018interiornet}, CO3D \cite{reizenstein2021common}, and 7-Scenes \cite{glocker2013real} datasets. We report the PSNR, MAE, and LPIPS metrics averaged over 1000 sampled images. The codebooks were evaluated with image size $128 \times 128$, except for `CO3D@400', which was evaluated with image size $400 \times 400$ pixels\label{tab:codebook-all}}
\setlength\tabcolsep{0.05cm}
\small\centering
\begin{tabular}{
@{}
l@{\extracolsep{0.1cm}}
rrr@{\extracolsep{0.1cm}}
@{}}
dataset & PSNR$\uparrow$ & MAE$\downarrow$ & LPIPS$\downarrow$ \\
\cmidrule{1-1} \cmidrule{2-4}
SM7 & 36.96 & 1.06 & 0.0075 \\
InteriorNet & 24.86 & 11.01 & 0.1966 \\
CO3D & 25.14 & 5.70 & 0.0994 \\
CO3D@400 & 25.34 & 5.63 & 0.1670 \\
7-Scenes (fine-tuned) & 19.29 & 17.51 & 0.2937 \\
7-Scenes & 19.00 & 19.22 & 0.3621 \\
ShapeNet-cars & 23.50 & 5.46 & 0.0734 \\
ShapeNet-chairs & 27.43 & 2.75 & 0.0425 \\
\bottomrule
\end{tabular}
\end{table}
In this section, we add more details on the codebook's representation capabilities (see \cref{fig:codebook} in the main paper) by showing quantitative results.
We evaluated the codebook models on each dataset's test set.
We report the peak signal-to-noise ratio (PSNR), mean absolute error computed in the RGB image space (MAE), and the LPIPS distance \cite{zhang2018unreasonable}. All codebooks were evaluated with image size $128\times128$ pixels except for `CO3D@400', which was evaluated with image size $400\times400$ pixels to be comparable with \cite{reizenstein2021common}. 
The metrics are averaged over 1000 randomly sampled images. 
The results can be seen in \cref{tab:codebook-all}.

Before training the final codebook, we experimented with different codebook models.
We also trained the DALL·E codebook \cite{ramesh2021zero}, which yielded slightly blurry images even when we used a codebook of size 8192 (normally, we use a codebook of size 1024). We observed a similar outcome with our codebook when we did not use the perceptual loss. We also tried to use a GAN loss for the codebook, as described in \cite{esser2021taming}. 
However, the generated images did not look geometrically consistent.

\section{Training details}\label{sec:training_details}
To allow our results to be reproduced, we give the details on the architecture of our method as well as the training hyperparameters.

All our \textbf{codebook models} were trained using the same set of hyperparameters.\footnote{Except for the SM7 dataset, where we only fine-tuned an existing model.} We trained codebooks of size 1024. 
The architecture is very similar to~\cite{esser2021taming} and is summarized in \cref{sec:codebook_architecture}.
We used the Adam optimizer~\cite{kingma2015adam} with learning rate\footnote{The learning rate was rescaled from prior experiments; $1.6 \times 10^{-3}$ would work too.} $1.584 \times 10^{-3}$ to train for 200k steps (roughly 480~GPU-hours) with a batch size of 352. 
For the CO3D dataset, we trained on the same 10 object categories as in \cite{reizenstein2021common} as well as on the full dataset. For the 7-Scenes dataset, due to not having enough images to train from scratch, we finetuned an InteriorNet pre-trained model. Therefore, we used only 20k batch updates with the same hyperparameters.

The architecture of our \textbf{transformer model} is based on GPT2-base \cite{radford2019language}, and has 12 transformer blocks, 12 attention heads, and the hidden size is 768. The model design was chosen based on its successes in other domains and because its size fits well on our hardware.
We trained our transformer models using the AdamW optimizer \cite{loshchilov2018decoupled}; we used the cosine schedule for the learning rate with a 2k step linear warmup.

For the \textbf{InteriorNet dataset}, we used the mixed-precision training with learning rate $8 \times 10^{-5}$, batch size 40, and learning rate decay $0.01$. The context size was 19, but we did not optimize the first four views. The weight of the localization loss term was 5. In all other experiments, the localization loss weight was 1 unless stated otherwise. 

For the \textbf{Shepard-Metzler-7-Parts (SM7) \cite{eslami2018neural,shepard1971mental}} dataset, we trained the transformer for 120k steps with the context size 5, batch size 128, and the learning rate $10^{-4}$ (cosine decay, warmup). Before passing camera poses into the transformer, we normalized the positions by multiplying them by $0.2$. We also gradually increased the weight of the localization term from 0 to 1 using the cosine schedule.\footnote{The schedule is not needed for the training to work and in newer experiments, we use a constant instead.}

For the \textbf{CO3D dataset}, we fine-tuned the model trained on the InteriorNet dataset. For the 10 categories, we optimized the model for 40k gradient steps with learning rate $10^{-4}$ (cosine decayed with a 2,000 step warmup), weight decay $0.05$, and batch size 80, employing mixed-precision training. The context size was 9, and the batch size was 80. We scaled the camera positions by $0.05$ in order for the positions to have a similar range as the pre-trained model. We also trained a model on all dataset categories using 100k gradient steps with the batch size 40, without using mixed-precision training, and when using the localization, we further used gradient clipping with the norm $1$ to improve stability.

For the \textbf{7-Scenes dataset}, we used a single InteriorNet pre-trained model which we fine-tuned on all 7-Scenes scenes. Same as in the original model, the context size was 19, but we did not optimize the first four views.  The transformer was fine-tuned for 10k gradient steps with learning rate $10^{-5}$ (cosine schedule, warmup). We rescaled the positions by multiplying them by $5$ to be in the same range as InteriorNet. 

Finally, for the \textbf{ShapeNet dataset}, we fine-tuned InteriorNet pre-trained model as well. We trained a single model for both categories: cars and chairs with the context size 3. We did not use mixed-precision training and the batch size was 64. The transformer was fine-tuned for 100k gradient steps with learning rate $10^{-4}$ (cosine schedule, warmup), weight decay was $0.05$, and we used gradient clipping with the norm $1$.

\section{Codebook architecture}\label{sec:codebook_architecture}
In \cref{tab:codebook} we give the details on the codebook architecture (\cf \cref{sec:method} in the main paper). 
The codebook model architecture was taken from \cite{esser2021taming} and modified slightly to downscale the images into two times smaller latent space. We have chosen this architecture because it had shown promising results for image generation in combination with transformers \cite{esser2021taming}. The other architecture we had considered was DALL·E \cite{ramesh2021zero}, but from our experiments, it performed worse.
\begin{table}[t]
\caption{
Codebook architecture details: the encoder (\textbf{top left}), the decoder (\textbf{right}), and the residual block (\textbf{bottom left}). For each layer, we list the number of output features (Num. features) and their sizes (Out. size). We denote kernel size as `ks', stride as `s', and the number of groups as `g'.
 We use nearest neighbor for the Upsample 2D layer.
 Note that the output of the residual block is added to its input as in ResNets \cite{he2016deep}. If the number of input channels is not equal to the number of output channels, the residual connection is implemented by applying an affine transformation to the input features position-wise before summing them with the output of this block
\label{tab:codebook}}
\tiny\centering
\newcommand\hcline{\cmidrule{1-1}\cmidrule{2-3}}
\begin{minipage}{0.47\linewidth}
\centering
\subfloat[Encoder]{
\begin{tabular}{@{}
l@{\extracolsep{0.1cm}}
rr@{}}
Layer type            & Num. features  & Out. size   \\
\hcline
Conv 2D (ks: 3)       & 128            & 128         \\
\hcline
ResBlock              & 128            & 128         \\
ResBlock              & 128            & 128         \\
Conv 2D (ks: 3, s: 2) & 128            & 64          \\
\hcline
ResBlock              & 128            & 64          \\
ResBlock              & 128            & 64          \\
Conv 2D (ks: 3, s: 2) & 128            & 32          \\
\hcline
ResBlock              & 256            & 32          \\
ResBlock              & 256            & 32          \\
Conv 2D (ks: 3, s: 2) & 256            & 16          \\
\hcline
ResBlock              & 256            & 16          \\
Attention 2D          & 256            & 16          \\
ResBlock              & 256            & 16          \\
Attention 2D          & 256            & 16          \\
Conv 2D (ks: 3, s: 2) & 256            & 8           \\
\hcline
ResBlock              & 512            & 8           \\
ResBlock              & 512            & 8           \\
\hcline
ResBlock              & 512            & 8           \\
Attention 2D          & 512            & 8           \\
ResBlock              & 512            & 8           \\
\hcline
GroupNorm 2D \cite{wu2018group} (g: 32)  & 512            & 8           \\
Swish \cite{ramachandran2017searching}    & 512            & 8           \\
Conv 2D (ks: 3)       & 256            & 8           \\
Conv 2D (ks: 1)       & 256            & 8           \\
\hline
\end{tabular}}\\
\subfloat[ResBlock]{
\begin{tabular}{
@{}
l@{\extracolsep{0.1cm}}
r
@{}
}
    Layer & Num. features \\
    \cmidrule{1-1} \cmidrule{2-2}
    GroupNorm \cite{wu2018group} (g: 32) & in \\
    Swish \cite{ramachandran2017searching} & in \\
    Conv 2D (ks: 3)   & out \\
    GroupNorm \cite{wu2018group} (g: 32) & out \\
    Swish \cite{ramachandran2017searching} & out \\
    Conv 2D (ks: 3)   & out \\
    \hline
\end{tabular}
}
\end{minipage}
\quad
\begin{minipage}{0.48\linewidth}
\centering
\subfloat[Decoder]{
\begin{tabular}{@{}
l@{\extracolsep{0.1cm}}
rr@{}}
Layer type            & Num. features  & Out. size   \\
\hcline
Conv 2D (ks: 1)       & 256            & 8           \\
Conv 2D (ks: 3)       & 512            & 8           \\
\hcline
ResBlock              & 512            & 8           \\
Attention 2D          & 512            & 8           \\
ResBlock              & 512            & 8           \\
\hcline
ResBlock              & 512            & 8           \\
ResBlock              & 512            & 8           \\
ResBlock              & 512            & 8           \\
Upsample 2D           & 512            & 16          \\
Conv 2D (ks: 3)       & 512            & 16          \\
\hcline
ResBlock              & 256            & 16          \\
Attention 2D          & 256            & 16          \\
ResBlock              & 256            & 16          \\
Attention 2D          & 256            & 16          \\
ResBlock              & 256            & 16          \\
Attention 2D          & 256            & 16          \\
Upsample 2D           & 256            & 32          \\
Conv 2D (ks: 3)       & 256            & 32          \\
\hcline
ResBlock              & 256            & 32          \\
ResBlock              & 256            & 32          \\
ResBlock              & 256            & 32          \\
Upsample 2D           & 256            & 64          \\
Conv 2D (ks: 3)       & 256            & 64          \\
\hcline
ResBlock              & 128            & 64          \\
ResBlock              & 128            & 64          \\
ResBlock              & 128            & 64          \\
Upsample 2D           & 128            & 128         \\
Conv 2D (ks: 3)       & 128            & 128         \\
\hcline
ResBlock              & 128            & 128         \\
ResBlock              & 128            & 128         \\
ResBlock              & 128            & 128         \\
\hcline
GroupNorm 2D \cite{wu2018group} (g: 32)  & 128            & 128           \\
Swish \cite{ramachandran2017searching}    & 128            & 128         \\
Conv 2D (ks: 3)       & 128            & 3           \\
\hline
\end{tabular}
}
\end{minipage}
\end{table}

\end{document}